\documentclass[letterpaper]{article} 
\usepackage[preprint]{aaai2027}
\usepackage[hyphens]{url}  
\usepackage{graphicx} 
\usepackage{natbib}  
\usepackage{caption} 
\usepackage{amsmath}
\usepackage{amssymb}
\usepackage{bm}
\usepackage{booktabs}
\usepackage{multirow}
\usepackage{colortbl}
\usepackage{placeins}

\definecolor{TableBackbone}{HTML}{E6E8E8}
\definecolor{TableField}{HTML}{DCE8ED}
\definecolor{TableForce}{HTML}{ECE2D4}
\definecolor{TableSummary}{HTML}{D7E7E3}
\definecolor{TableAverage}{HTML}{E6E8E8}
\definecolor{TableBackboneBody}{HTML}{F0F1F1}
\definecolor{TableFieldBody}{HTML}{EDF3F5}
\definecolor{TableForceBody}{HTML}{F4EFE8}
\definecolor{TableSummaryBody}{HTML}{EBF3F1}
\definecolor{STCOAccent}{HTML}{B7473A}
\newcommand{\STCOResult}[1]{{\color{STCOAccent}#1}}
\makeatletter
\newcommand{\TableEdgeCell}[3]{%
  \gdef\CT@cell@color{%
    \CT@color{#1}%
    \@tempdimb#2\relax
    \@tempdimc#3\relax
    \global\let\CT@cell@color\relax}}
\makeatother

\graphicspath{{figures/}}

\newcommand{\STCOMeanFieldGain}{31.1\%}
\newcommand{\STCOMeanLoadGain}{24.7\%}
\newcommand{\STCOFieldWins}{68}
\newcommand{\STCOLoadWins}{63}
\newcommand{\STCOStrata}{72}
\newcommand{\STCOAggregateWins}{12}
\newcommand{\STCOOODWins}{11}
\newcommand{\STCOObservedContextWins}{11}
\newcommand{\STCOTimeContextWins}{5}
\newcommand{\STCOLagIDGain}{53.4\%}
\newcommand{\STCOLagOODGain}{45.6\%}

\title{STCO: Conditional Neural Operators for Time-Dependent PDEs}
\author{Xingxin Yang, Zhan Zhang, Juan Li$^{*}$}
\affiliations{Department of Engineering, King's College London, London, United Kingdom\\
$^{*}$Correspondence: juan.li@kcl.ac.uk}

\begin{document}

\maketitle

\begin{abstract}

Neural operators have emerged as efficient surrogates for time-dependent physical systems governed by partial differential equations (PDEs),
but their future-state predictions are often conditioned only on observed states and static problem descriptors.
For control or optimization, however, body motion, inflow, or forcing are prescribed for the query without being determined solely by the observed state.
We introduce the Spatiotemporal Conditional Operator (STCO) for prescribed-condition operator learning (PCOL),
a common interface that supplies prescribed target-time condition fields to heterogeneous backbone architectures while retaining their architecture-specific core computation and context pathways.
Its condition interface combines Flow-Aware Graph Leaf (FAGL) with Dual-Site Feature-wise Linear Modulation (DSFiLM).
Non-learned FAGL uses vorticity from the final observed frame to construct a fixed-cardinality adaptive partition,
then co-locates the observed history and target-time condition fields at its regional coordinates.
DSFiLM injects separate motion, inflow, and force routes before and after operator computation through current-feature-driven slot- and channel-wise gates.
We evaluate twelve matched backbone architectures with different existing physical and temporal inputs.
The immersed-boundary computational fluid dynamics (CFD) benchmark spans prescribed motion, inflow disturbances, body-force actuation, and morphology.
Across twelve matched backbones, three regimes, and two lead ranges, STCO yields mean paired reductions of \STCOMeanFieldGain{} in relative-$L_2$ field error and \STCOMeanLoadGain{} in normalized pressure-derived load error.
It also lowers longer-lead field error for \STCOOODWins{} backbones, while interventions on individual condition groups produce measurable prediction changes for every group evaluated.

\end{abstract}

\section{Introduction}

Neural operators learn mappings between function spaces and provide efficient surrogates for parametric partial differential equations (PDEs)
\cite{li2020fno,lu2021deeponet,li2023gino}.
For time-dependent PDEs, they commonly forecast a future field from an initial state or observed trajectory, with lead time or auxiliary trajectories supplied as context
\cite{alkin2024upt,herde2024poseidon,mousavi2025rigno,koupai2025enma}.
This formulation is effective when the requested future is determined by the observed dynamics and fixed problem descriptors.

\begin{figure*}[t]
    \centering
    \includegraphics[width=\textwidth]{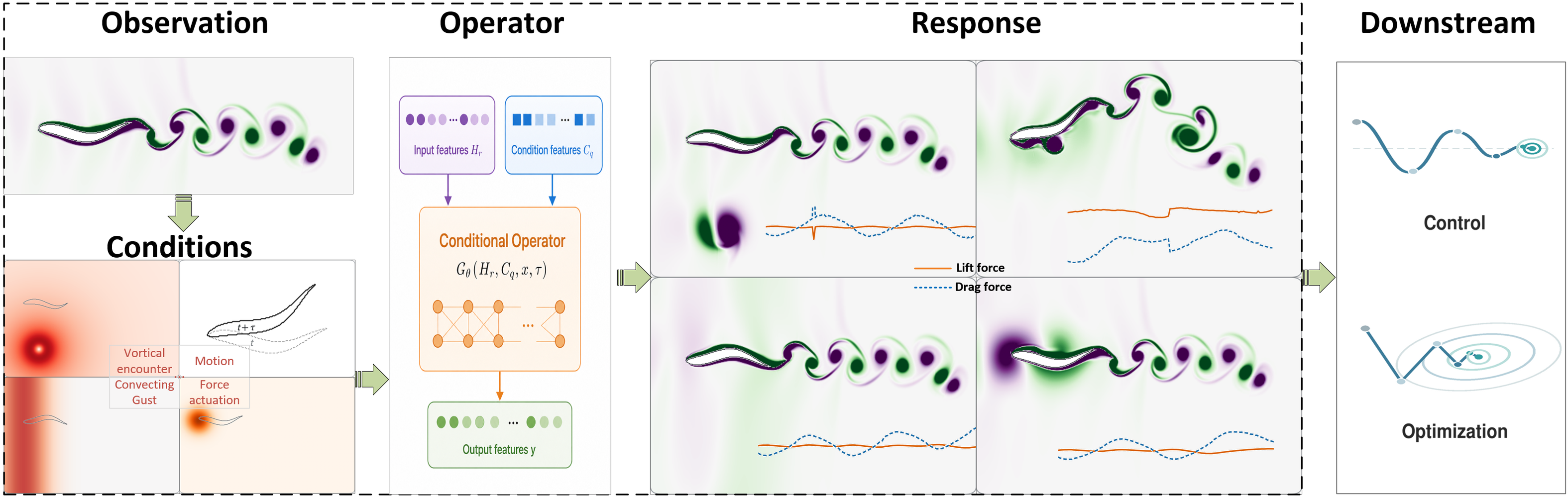}
    \caption{Prescribed-condition operator learning in moving-body flow. The dashed frame marks the PCOL map from observed history and prescribed condition to future response. Examples include a vortical encounter, body motion, a convecting gust, and body-force actuation. Response panels show vorticity and pressure-derived load histories from CFD ground truth (GT). The predicted velocity--pressure field supports downstream control and optimization.}
    \label{fig:prescribed-condition-pde}
\end{figure*}

Control and optimization often require a different query.
The system response must be evaluated under a future scenario specified for the query, such as a gust, target body geometry, or prescribed boundary input
\cite{bartels2013gustmodel,sedky2022gust,han2021movingboundary,hu2025safepdecontrol}.
The history describes the current dynamics, while the prescribed condition distinguishes the future to be evaluated.
As shown in Figure~\ref{fig:prescribed-condition-pde}, we formulate this prediction problem as \emph{prescribed-condition operator learning} (PCOL).
In PCOL, a conditional neural operator maps an observed history, a lead time, and physical condition fields specified at the target time to the corresponding future response.
This query structure is also shared by predictive world models that infer future states from past observations and a supplied action or intervention
\cite{agarwal2025cosmos,hafner2025dreamer}.
PCOL resolves the response as a PDE solution field that can be evaluated in downstream control or optimization.

We examine PCOL through two-dimensional incompressible moving-body flow.
Geometry and motion alter the boundary, inflow disturbances modify inlet data, and distributed forces enter the momentum equation, allowing distinct prescribed mechanisms to be varied within one Navier--Stokes system
\cite{mittal2005ib,sotiropoulos2014immersed}.
This setting exposes three linked challenges.
\textbf{1) Representation alignment.}
Observed and prescribed fields differ in channels, spatial support, and physical meaning.
\textbf{2) Selective condition modulation.}
The model must preserve the roles of geometry, inflow, and forcing while adapting their influence to the current features.
\textbf{3) Sparse-event supervision.}
Brief force and inflow events provide few condition-revealing pairs under uniform temporal sampling.

We introduce the \emph{Spatiotemporal Conditional Operator} (STCO), shown in Figure~\ref{fig:stco-overview}, as a common condition interface for heterogeneous backbone architectures.
Its \emph{Flow-Aware Graph Leaf} (FAGL) representation forms a fixed-cardinality, vorticity-aware partition and co-locates historical and prescribed fields at shared regional slots.
\emph{Dual-Site Feature-wise Linear Modulation} (DSFiLM) assigns framewise motion, inflow, and force to separate routes, with signed-distance geometry localizing the motion route.
Its independently parameterized IN-DSFiLM and OUT-DSFiLM modules act before and after the backbone core, respectively, while current features gate each route by slot and channel.
External-activity sampling increases exposure to onset crossings and active intervals.
STCO adds target-time conditioning while retaining each backbone's core computation and existing context pathways.

We evaluate this interface through twelve matched pairs of baseline (Base) and STCO configurations.
Within each pair, FAGL, external-activity sampling, data, backbone, decoder, and training protocol are fixed.
The Base models retain their original inputs.
Eleven receive observed-frame conditions, and five also encode lead time.
STCO preserves these inputs and activates DSFiLM, which combines target-time condition routes with lead-time modulation.
The paired differences measure the predictive value of this complete interface under an otherwise matched representation and training protocol.
STCO yields a positive mean field gain across the six regime--lead strata for all \STCOAggregateWins{} backbones, lowers the field error beyond the training lead range for \STCOOODWins{}, and reduces pressure-derived load error overall.
\emph{Modulation counterfactual} (MCF) interventions establish prediction sensitivity to every evaluated spatial condition group.

Our contributions are summarized below.
\begin{itemize}
    \item We formulate PCOL as predicting a future response from observed history and physical conditions prescribed for the query. We address this problem with STCO, a common condition interface that retains the core computation of heterogeneous backbone architectures.
    \item We introduce vorticity-aware FAGL for adaptive spatial alignment of observed fields and prescribed conditions. DSFiLM modulates separate motion, inflow, and force routes before and after backbone computation, with their influence gated by the current features.
    \item We construct a moving-body-flow benchmark and evaluate STCO in matched comparisons across twelve backbones. External-activity sampling increases exposure to sparse force and inflow events. MCF measures prediction sensitivity to each evaluated spatial condition group.
\end{itemize}

\section{Related Work}

\textbf{Prescribed response and conditional operators.}
Moving-boundary predictors infer the next flow field from recent flow fields and boundary-position histories
\cite{han2021movingboundary}.
A closely related rigid-body fluid--structure interaction study predicts the terminal fluid state from an initial state and a prescribed motion sequence over the full prediction interval
\cite{zhong2025fsi}.
It compares direct concatenation, temporal pooling, and sequential coupling.
Boundary-control operators map temporal boundary inputs to boundary-output trajectories
\cite{hu2025safepdecontrol}.
Other neural operators evaluate candidate controls within an outer optimization loop or predict controller gains and feedback inputs
\cite{hwang2022pdecontrol,bhan2024backstepping}.
PCOL addresses forward response from observed history and target-time condition fields, leaving control selection downstream.
MIONet represents products of function spaces
\cite{jin2022mionet}, while GNOT, GINO, and Unisolver process heterogeneous inputs, geometry, or structured PDE descriptors
\cite{hao2023gnot,li2023gino,zhou2024unisolver}.
GEPS learns environment conditioning, and time-dependent operators incorporate lead time or context trajectories
\cite{koupai2024geps,herde2024poseidon,mousavi2025rigno,koupai2025enma}.
STCO organizes multiple target-time physical fields through one interface rather than introducing another backbone architecture.

\textbf{Adaptive representation and modulation.}
Irregular-domain operators use coordinate deformation, graph--grid transfer, regional graphs, or latent tokens
\cite{li2023geofno,li2023gino,mousavi2025rigno,alkin2024upt}.
UPT also conditions Transformer and Perceiver blocks on time or velocity context through feature modulation
\cite{alkin2024upt}.
FAGL draws on flow-guided refinement
\cite{kamkar2011feature} and graph-based component merging
\cite{felzenszwalb2004graphseg} to allocate a fixed slot budget and align fields at shared indices.
FiLM applies condition-dependent affine transformations, SPADE makes them spatially varying, and squeeze--excitation derives channel weights from current features
\cite{perez2018film,park2019spade,hu2018senet}.
DSFiLM combines these principles through separate spatial condition routes, local state-dependent gates, and independently learned IN-DSFiLM and OUT-DSFiLM modules.

\section{Methodology}
\label{sec:methodology}

\subsection{Problem Formulation}

\textbf{Setup.}
We instantiate PCOL for time-dependent PDEs in incompressible moving-body flow.
Let $\bm x\in\Omega(t)\subset\mathbb R^2$ denote a point in the time-dependent fluid domain and
$\partial\Omega_b(t)$ the moving body boundary.
Using body chord $L$, reference inflow speed $U_\infty$, and constant density $\rho$ as the length, velocity, and density scales, we scale time, pressure, and acceleration by
$L/U_\infty$, $\rho U_\infty^2/2$, and $U_\infty^2/L$, respectively.
After dropping dimensionless superscripts, the velocity
$\bm u=(u,v)$ and pressure coefficient $p$ satisfy
\begin{equation}
\begin{aligned}
    \partial_t\bm u+(\bm u\cdot\nabla)\bm u
        &=-\tfrac12\nabla p
          +\mathrm{Re}^{-1}\nabla^2\bm u+\bm g,\\
    \nabla\cdot\bm u&=0 .
\end{aligned}
    \label{eq:moving-body-ns}
\end{equation}
Here, $\mathrm{Re}=U_\infty L/\nu$, $\nu$ is the kinematic viscosity, and
$\bm g(\bm x,t)$ is a prescribed body-force acceleration.
The prescribed boundary conditions are $\bm u=\bm u_b$ on
$\partial\Omega_b(t)$ and
$\bm u=\bm e_x+\bm u_{\mathrm{bc}}$ on the inlet
$\Gamma_{\mathrm{in}}$.
Here, $\bm u_b$ is the prescribed body velocity,
$\bm e_x=(1,0)^\top$ is the nondimensional base inflow, and
$\bm u_{\mathrm{bc}}$ is the prescribed inflow perturbation.
Body kinematics define the signed-distance field $\psi(\bm x,t)$, whose zero level set is $\partial\Omega_b(t)$.
We sample $T$ stored frames at times $\{t_n\}_{n=0}^{T-1}$.
At frame $n$, the domain is $\Omega_n=\Omega(t_n)$, and for $\bm x\in\Omega_n$ we define the response and prescribed-condition fields
\begin{equation}
\begin{aligned}
    \bm y_n(\bm x)
        &= [u_n(\bm x),v_n(\bm x),p_n(\bm x)]^\top,\\
    \bm c_n(\bm x)
        &= [\psi_n(\bm x),\Delta\psi_n(\bm x),
            \bm g_n(\bm x)^\top,
            \bm u_{\mathrm{bc},n}(\bm x)^\top]^\top .
\end{aligned}
\label{eq:pcol-fields}
\end{equation}
The framewise signed-distance change is
$\Delta\psi_n=\psi_n-\psi_{n-1}$ for $n\geq1$, with
$\Delta\psi_0=0$.
The six condition channels encode geometry and framewise motion through $\psi_n$ and $\Delta\psi_n$, distributed forcing through $\bm g_n$, and inflow perturbation through $\bm u_{\mathrm{bc},n}$.

\textbf{Objective.}
Let $n_r$ be the final observed reference index and $n_q>n_r$ the queried future index.
Subscripts $r$ and $q$ henceforth denote evaluation at $n_r$ and $n_q$.
The observed history is
$\mathcal H_r=\{(\bm y_{n_k},\bm c_{n_k})\}_{k=1}^{K}$, where
$K$ is the number of observed frames and
$n_1<\cdots<n_K=n_r$.
The query lead is $\Delta n=n_q-n_r$, represented by the normalized scalar
$\tau=\Delta n/(T-1)$.
PCOL seeks an operator that maps the observed history and the condition prescribed for this query to its future response
\begin{equation}
    \widehat{\bm y}_q(\bm x)
    =\mathcal G_\theta
      \bigl(\mathcal H_r,\bm c_q,\bm x,\tau\bigr)
    \approx \bm y_q(\bm x),
    \qquad \bm x\in\Omega_q ,
    \label{eq:pcol-objective-map}
\end{equation}
where $\bm y_q=\bm y_{n_q}$, $\bm c_q=\bm c_{n_q}$,
$\Omega_q=\Omega_{n_q}$, and $\theta$ contains the learned parameters.

\subsection{STCO Architecture}

\begin{figure*}[t]
    \centering
    \includegraphics[width=0.976\textwidth]{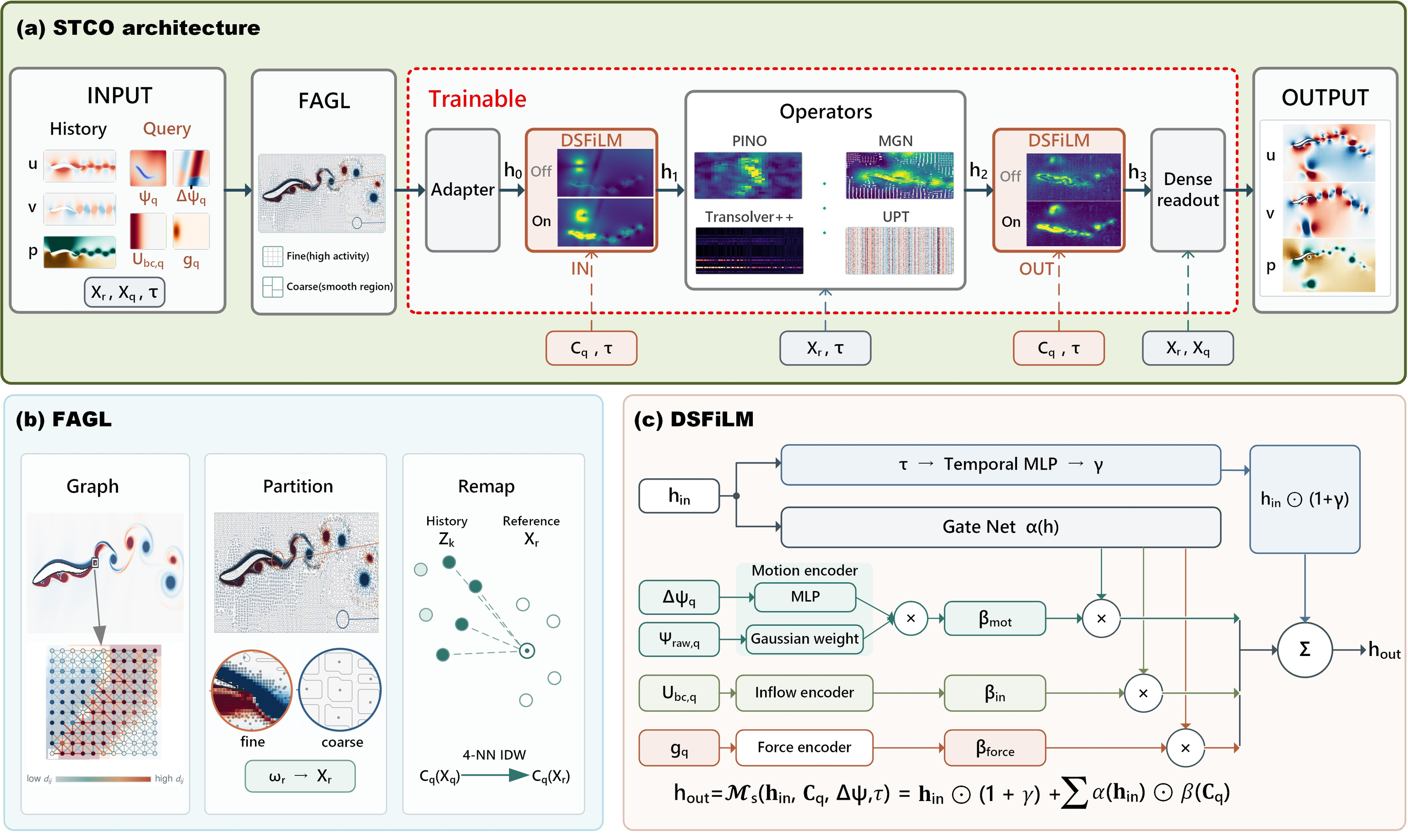}
    \caption{STCO architecture. (a) Overall pipeline from observed history and prescribed target-time conditions to the velocity--pressure response on $\bm X_q$. FAGL provides regional coordinates $\bm X_r$, while IN-DSFiLM and OUT-DSFiLM bracket a selected backbone core. (b) FAGL constructs an observed-vorticity-aware partition and aligns historical aggregates and target-time conditions on $\bm X_r$ by four-neighbor inverse-distance weighting. (c) DSFiLM applies lead-time scaling and state-gated motion, inflow, and force shifts, with signed distance localizing motion near the body.}
\label{fig:stco-overview}
\end{figure*}

\newcommand{\ResultsTableAndPerformanceFloat}{%
\begin{table*}[t!]
\centering
\small
\renewcommand{\STCOResult}[1]{##1}
\setlength{\tabcolsep}{2.5pt}
\begin{tabular}{@{}l*{15}{c}@{}}
\toprule
\TableEdgeCell{TableBackbone}{0pt}{\tabcolsep}Backbone
& \multicolumn{3}{c}{\TableEdgeCell{TableField}{\tabcolsep}{\tabcolsep}Field $E_y$, ID}
& \multicolumn{3}{c}{\TableEdgeCell{TableField}{\tabcolsep}{\tabcolsep}Field $E_y$, OOD}
& \multicolumn{3}{c}{\TableEdgeCell{TableForce}{\tabcolsep}{\tabcolsep}Load $E_{F_p}$, ID}
& \multicolumn{3}{c}{\TableEdgeCell{TableForce}{\tabcolsep}{\tabcolsep}Load $E_{F_p}$, OOD}
& \multicolumn{1}{c}{\TableEdgeCell{TableSummary}{\tabcolsep}{\tabcolsep}Field gain}
& \multicolumn{2}{c}{\TableEdgeCell{TableSummary}{\tabcolsep}{\tabcolsep}MCF sensitivity} \\
\TableEdgeCell{TableBackbone}{0pt}{\tabcolsep}
& \cellcolor{TableField}Base & \cellcolor{TableField}STCO
& \cellcolor{TableField}$\Delta$ (\%)
& \cellcolor{TableField}Base & \cellcolor{TableField}STCO
& \cellcolor{TableField}$\Delta$ (\%)
& \cellcolor{TableForce}Base & \cellcolor{TableForce}STCO
& \cellcolor{TableForce}$\Delta$ (\%)
& \cellcolor{TableForce}Base & \cellcolor{TableForce}STCO
& \cellcolor{TableForce}$\Delta$ (\%)
& \cellcolor{TableSummary}Mean $\Delta_y$ (\%)
& \cellcolor{TableSummary}Lead
& \TableEdgeCell{TableSummary}{\tabcolsep}{0pt}Spatial \\
\midrule
\TableEdgeCell{TableBackboneBody}{0pt}{\tabcolsep}MGN & \cellcolor{TableFieldBody}0.438 & \cellcolor{TableFieldBody}\STCOResult{\textbf{0.242}} & \cellcolor{TableFieldBody}$+44.8$ & \cellcolor{TableFieldBody}0.802 & \cellcolor{TableFieldBody}\STCOResult{\textbf{0.713}} & \cellcolor{TableFieldBody}$+11.1$ & \cellcolor{TableForceBody}0.772 & \cellcolor{TableForceBody}\STCOResult{\textbf{0.606}} & \cellcolor{TableForceBody}$+21.5$ & \cellcolor{TableForceBody}0.774 & \cellcolor{TableForceBody}\STCOResult{\textbf{0.613}} & \cellcolor{TableForceBody}$+20.8$ & \cellcolor{TableSummaryBody}$+26.8$ & \cellcolor{TableSummaryBody}0.386 & \TableEdgeCell{TableSummaryBody}{\tabcolsep}{0pt}0.293 \\
\TableEdgeCell{TableBackboneBody}{0pt}{\tabcolsep}RIGNO & \cellcolor{TableFieldBody}0.243 & \cellcolor{TableFieldBody}\STCOResult{\textbf{0.211}} & \cellcolor{TableFieldBody}$+13.3$ & \cellcolor{TableFieldBody}0.493 & \cellcolor{TableFieldBody}\STCOResult{\textbf{0.445}} & \cellcolor{TableFieldBody}$+9.8$ & \cellcolor{TableForceBody}0.597 & \cellcolor{TableForceBody}\STCOResult{\textbf{0.592}} & \cellcolor{TableForceBody}$+0.9$ & \cellcolor{TableForceBody}\textbf{0.564} & \cellcolor{TableForceBody}\STCOResult{0.606} & \cellcolor{TableForceBody}$-7.6$ & \cellcolor{TableSummaryBody}$+11.4$ & \cellcolor{TableSummaryBody}0.318 & \TableEdgeCell{TableSummaryBody}{\tabcolsep}{0pt}0.352 \\
\TableEdgeCell{TableBackboneBody}{0pt}{\tabcolsep}PINO & \cellcolor{TableFieldBody}0.438 & \cellcolor{TableFieldBody}\STCOResult{\textbf{0.194}} & \cellcolor{TableFieldBody}$+55.7$ & \cellcolor{TableFieldBody}0.765 & \cellcolor{TableFieldBody}\STCOResult{\textbf{0.353}} & \cellcolor{TableFieldBody}$+53.9$ & \cellcolor{TableForceBody}0.879 & \cellcolor{TableForceBody}\STCOResult{\textbf{0.559}} & \cellcolor{TableForceBody}$+36.5$ & \cellcolor{TableForceBody}0.776 & \cellcolor{TableForceBody}\STCOResult{\textbf{0.437}} & \cellcolor{TableForceBody}$+43.6$ & \cellcolor{TableSummaryBody}$+53.6$ & \cellcolor{TableSummaryBody}0.211 & \TableEdgeCell{TableSummaryBody}{\tabcolsep}{0pt}0.374 \\
\TableEdgeCell{TableBackboneBody}{0pt}{\tabcolsep}Poseidon & \cellcolor{TableFieldBody}0.206 & \cellcolor{TableFieldBody}\STCOResult{\textbf{0.154}} & \cellcolor{TableFieldBody}$+25.4$ & \cellcolor{TableFieldBody}0.358 & \cellcolor{TableFieldBody}\STCOResult{\textbf{0.279}} & \cellcolor{TableFieldBody}$+22.1$ & \cellcolor{TableForceBody}0.572 & \cellcolor{TableForceBody}\STCOResult{\textbf{0.544}} & \cellcolor{TableForceBody}$+4.9$ & \cellcolor{TableForceBody}0.490 & \cellcolor{TableForceBody}\STCOResult{\textbf{0.434}} & \cellcolor{TableForceBody}$+11.5$ & \cellcolor{TableSummaryBody}$+24.2$ & \cellcolor{TableSummaryBody}0.293 & \TableEdgeCell{TableSummaryBody}{\tabcolsep}{0pt}0.323 \\
\TableEdgeCell{TableBackboneBody}{0pt}{\tabcolsep}GAOT & \cellcolor{TableFieldBody}0.431 & \cellcolor{TableFieldBody}\STCOResult{\textbf{0.139}} & \cellcolor{TableFieldBody}$+67.7$ & \cellcolor{TableFieldBody}0.786 & \cellcolor{TableFieldBody}\STCOResult{\textbf{0.331}} & \cellcolor{TableFieldBody}$+57.9$ & \cellcolor{TableForceBody}0.783 & \cellcolor{TableForceBody}\STCOResult{\textbf{0.521}} & \cellcolor{TableForceBody}$+33.4$ & \cellcolor{TableForceBody}0.811 & \cellcolor{TableForceBody}\STCOResult{\textbf{0.479}} & \cellcolor{TableForceBody}$+40.9$ & \cellcolor{TableSummaryBody}$+61.5$ & \cellcolor{TableSummaryBody}0.232 & \TableEdgeCell{TableSummaryBody}{\tabcolsep}{0pt}0.393 \\
\TableEdgeCell{TableBackboneBody}{0pt}{\tabcolsep}GINO & \cellcolor{TableFieldBody}0.310 & \cellcolor{TableFieldBody}\STCOResult{\textbf{0.273}} & \cellcolor{TableFieldBody}$+12.1$ & \cellcolor{TableFieldBody}0.605 & \cellcolor{TableFieldBody}\STCOResult{\textbf{0.455}} & \cellcolor{TableFieldBody}$+24.8$ & \cellcolor{TableForceBody}0.690 & \cellcolor{TableForceBody}\STCOResult{\textbf{0.636}} & \cellcolor{TableForceBody}$+7.8$ & \cellcolor{TableForceBody}0.538 & \cellcolor{TableForceBody}\STCOResult{\textbf{0.478}} & \cellcolor{TableForceBody}$+11.2$ & \cellcolor{TableSummaryBody}$+18.3$ & \cellcolor{TableSummaryBody}0.254 & \TableEdgeCell{TableSummaryBody}{\tabcolsep}{0pt}0.622 \\
\TableEdgeCell{TableBackboneBody}{0pt}{\tabcolsep}Transolver++ & \cellcolor{TableFieldBody}0.434 & \cellcolor{TableFieldBody}\STCOResult{\textbf{0.157}} & \cellcolor{TableFieldBody}$+63.7$ & \cellcolor{TableFieldBody}0.773 & \cellcolor{TableFieldBody}\STCOResult{\textbf{0.525}} & \cellcolor{TableFieldBody}$+32.0$ & \cellcolor{TableForceBody}0.791 & \cellcolor{TableForceBody}\STCOResult{\textbf{0.528}} & \cellcolor{TableForceBody}$+33.2$ & \cellcolor{TableForceBody}0.775 & \cellcolor{TableForceBody}\STCOResult{\textbf{0.571}} & \cellcolor{TableForceBody}$+26.4$ & \cellcolor{TableSummaryBody}$+46.9$ & \cellcolor{TableSummaryBody}0.030 & \TableEdgeCell{TableSummaryBody}{\tabcolsep}{0pt}0.484 \\
\TableEdgeCell{TableBackboneBody}{0pt}{\tabcolsep}Unisolver & \cellcolor{TableFieldBody}0.257 & \cellcolor{TableFieldBody}\STCOResult{\textbf{0.204}} & \cellcolor{TableFieldBody}$+20.6$ & \cellcolor{TableFieldBody}0.558 & \cellcolor{TableFieldBody}\STCOResult{\textbf{0.509}} & \cellcolor{TableFieldBody}$+8.9$ & \cellcolor{TableForceBody}0.672 & \cellcolor{TableForceBody}\STCOResult{\textbf{0.596}} & \cellcolor{TableForceBody}$+11.3$ & \cellcolor{TableForceBody}0.615 & \cellcolor{TableForceBody}\STCOResult{\textbf{0.480}} & \cellcolor{TableForceBody}$+22.1$ & \cellcolor{TableSummaryBody}$+14.4$ & \cellcolor{TableSummaryBody}0.272 & \TableEdgeCell{TableSummaryBody}{\tabcolsep}{0pt}0.516 \\
\TableEdgeCell{TableBackboneBody}{0pt}{\tabcolsep}MPP & \cellcolor{TableFieldBody}0.466 & \cellcolor{TableFieldBody}\STCOResult{\textbf{0.247}} & \cellcolor{TableFieldBody}$+47.0$ & \cellcolor{TableFieldBody}0.781 & \cellcolor{TableFieldBody}\STCOResult{\textbf{0.584}} & \cellcolor{TableFieldBody}$+25.3$ & \cellcolor{TableForceBody}0.817 & \cellcolor{TableForceBody}\STCOResult{\textbf{0.612}} & \cellcolor{TableForceBody}$+25.1$ & \cellcolor{TableForceBody}0.747 & \cellcolor{TableForceBody}\STCOResult{\textbf{0.562}} & \cellcolor{TableForceBody}$+24.7$ & \cellcolor{TableSummaryBody}$+35.3$ & \cellcolor{TableSummaryBody}0.141 & \TableEdgeCell{TableSummaryBody}{\tabcolsep}{0pt}0.445 \\
\TableEdgeCell{TableBackboneBody}{0pt}{\tabcolsep}CALM-PDE & \cellcolor{TableFieldBody}0.191 & \cellcolor{TableFieldBody}\STCOResult{\textbf{0.135}} & \cellcolor{TableFieldBody}$+29.4$ & \cellcolor{TableFieldBody}\textbf{0.383} & \cellcolor{TableFieldBody}\STCOResult{0.447} & \cellcolor{TableFieldBody}$-16.8$ & \cellcolor{TableForceBody}0.522 & \cellcolor{TableForceBody}\STCOResult{\textbf{0.496}} & \cellcolor{TableForceBody}$+4.9$ & \cellcolor{TableForceBody}0.488 & \cellcolor{TableForceBody}\STCOResult{\textbf{0.463}} & \cellcolor{TableForceBody}$+4.9$ & \cellcolor{TableSummaryBody}$+5.1$ & \cellcolor{TableSummaryBody}0.019 & \TableEdgeCell{TableSummaryBody}{\tabcolsep}{0pt}0.413 \\
\TableEdgeCell{TableBackboneBody}{0pt}{\tabcolsep}UPT & \cellcolor{TableFieldBody}0.426 & \cellcolor{TableFieldBody}\STCOResult{\textbf{0.136}} & \cellcolor{TableFieldBody}$+68.2$ & \cellcolor{TableFieldBody}0.775 & \cellcolor{TableFieldBody}\STCOResult{\textbf{0.522}} & \cellcolor{TableFieldBody}$+32.6$ & \cellcolor{TableForceBody}0.775 & \cellcolor{TableForceBody}\STCOResult{\textbf{0.490}} & \cellcolor{TableForceBody}$+36.7$ & \cellcolor{TableForceBody}0.756 & \cellcolor{TableForceBody}\STCOResult{\textbf{0.530}} & \cellcolor{TableForceBody}$+29.9$ & \cellcolor{TableSummaryBody}$+49.6$ & \cellcolor{TableSummaryBody}0.010 & \TableEdgeCell{TableSummaryBody}{\tabcolsep}{0pt}0.419 \\
\TableEdgeCell{TableBackboneBody}{0pt}{\tabcolsep}GEPS & \cellcolor{TableFieldBody}0.475 & \cellcolor{TableFieldBody}\STCOResult{\textbf{0.300}} & \cellcolor{TableFieldBody}$+36.8$ & \cellcolor{TableFieldBody}0.807 & \cellcolor{TableFieldBody}\STCOResult{\textbf{0.680}} & \cellcolor{TableFieldBody}$+15.8$ & \cellcolor{TableForceBody}0.810 & \cellcolor{TableForceBody}\STCOResult{\textbf{0.682}} & \cellcolor{TableForceBody}$+15.8$ & \cellcolor{TableForceBody}0.783 & \cellcolor{TableForceBody}\STCOResult{\textbf{0.651}} & \cellcolor{TableForceBody}$+16.9$ & \cellcolor{TableSummaryBody}$+25.6$ & \cellcolor{TableSummaryBody}0.356 & \TableEdgeCell{TableSummaryBody}{\tabcolsep}{0pt}0.391 \\
\midrule
\TableEdgeCell{TableAverage}{0pt}{\tabcolsep}\textbf{Average} & \cellcolor{TableAverage}0.360 & \cellcolor{TableAverage}\STCOResult{\textbf{0.199}} & \cellcolor{TableAverage}$+40.4$ & \cellcolor{TableAverage}0.657 & \cellcolor{TableAverage}\STCOResult{\textbf{0.487}} & \cellcolor{TableAverage}$+23.1$ & \cellcolor{TableAverage}0.723 & \cellcolor{TableAverage}\STCOResult{\textbf{0.572}} & \cellcolor{TableAverage}$+19.3$ & \cellcolor{TableAverage}0.676 & \cellcolor{TableAverage}\STCOResult{\textbf{0.525}} & \cellcolor{TableAverage}$+20.4$ & \cellcolor{TableAverage}$+31.1$ & \cellcolor{TableAverage}0.210 & \TableEdgeCell{TableAverage}{\tabcolsep}{0pt}0.419 \\
\bottomrule
\end{tabular}
\caption{Regime-balanced accuracy and MCF sensitivity across twelve matched backbones. ID/OOD denote $\Delta n=1$--$20$/$21$--$40$; errors macro-average three regimes. Positive $\Delta$ favors STCO, and bold marks the lower paired error. The Average row macro-averages backbone-level entries, including paired reductions. Mean $\Delta_y$ averages six field gains; MCF reports lead-time and mean spatial-condition sensitivity.}
\label{tab:main-results}
\normalsize
\centering
\includegraphics[width=0.90\textwidth]{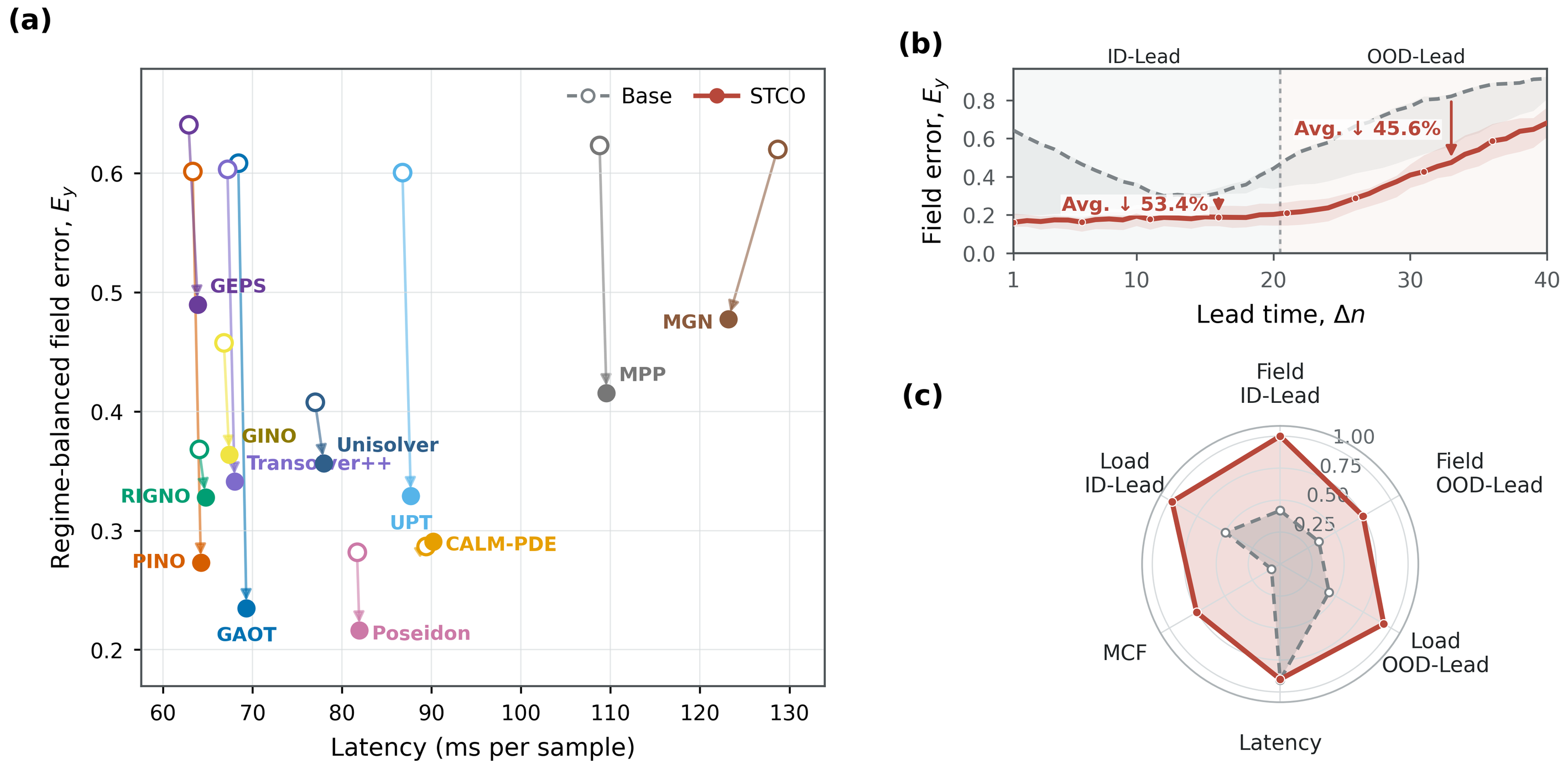}
\captionof{figure}{Cross-backbone accuracy, condition dependence, and inference cost. \textbf{(a)} H200 median dense-query latency at $N_q=264{,}196$ and batch size one versus six-stratum regime-balanced $E_y$; arrows connect matched Base and STCO configurations. \textbf{(b)} Cross-backbone median and interquartile range of $E_y$ at each lead. Annotations average the relative reductions of exact-lead medians within ID-Lead and OOD-Lead, whereas Mean $\Delta_y$ in Table~\ref{tab:main-results} averages six stratum-level gains per backbone. \textbf{(c)} Cross-backbone means of ID/OOD $E_y$, $E_{F_p}$, MCF, and latency under shared normalization; MCF equally weights lead, moving-boundary, force, and inflow responses. Accuracy and MCF use the same fixed population of $5{,}004$ reference--query pairs, with MCF applying controlled input interventions to each pair. Outward denotes lower error or latency and stronger MCF; polygon area is not an aggregate score.}
\label{fig:performance-summary}
\end{table*}
}

\newcommand{\SpatialResultsFloat}{%
\begin{figure*}[t!]
    \centering
    \includegraphics[width=0.865\textwidth]{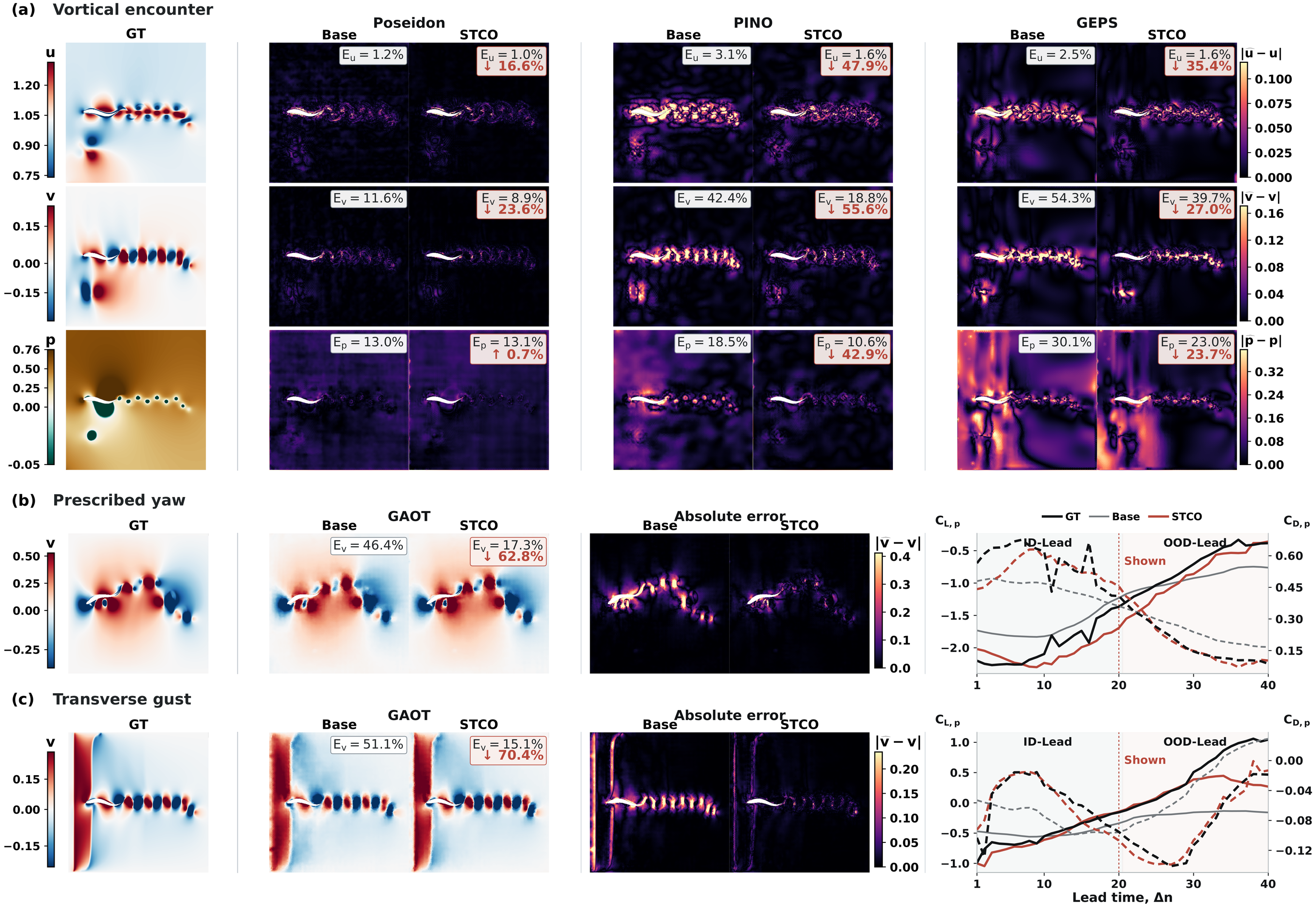}
    \caption{Prescribed-condition velocity--pressure responses and pressure-derived loads. \textbf{(a)} CFD GT and matched Base/STCO absolute errors for Poseidon, PINO, and GEPS on a vortical encounter at $n_r=128$, $\Delta n=10$; rows show $u$, $v$, and $p$. \textbf{(b,c)} CFD GT, GAOT Base/STCO predictions, absolute $v$ errors, and independent-query loads for prescribed yaw ($n_r=128$) and transverse gust ($n_r=126$) at $\Delta n=20$. Backgrounds mark ID/OOD-Lead, and vertical lines mark displayed fields. Annotations give relative-$L_2$ errors; color limits are shared within comparisons.}
    \label{fig:spatial-results}
\end{figure*}
}

STCO realizes the PCOL map through adaptive spatial alignment and conditional modulation, as shown in Figure~\ref{fig:stco-overview}(a).
FAGL co-locates the observed history and target-time condition fields at shared regional slots.
Backbone-specific adapters connect IN-DSFiLM and OUT-DSFiLM to the selected backbone core while preserving its computation.

Let $\bm X_r=[\bm x_{r,\ell}]_{\ell=1}^{N_\ell}\in\mathbb R^{N_\ell\times2}$
denote the $N_\ell$ final-observed-frame slot coordinates, and let
$\bm X_q=[\bm x_{q,j}]_{j=1}^{N_q}\in\mathbb R^{N_q\times2}$ contain the $N_q$ requested coordinates, with $\bm x_{q,j}\in\Omega_q$.
Let $\overline{\bm Z}_r\in\mathbb R^{K\times N_\ell\times9}$ contain the aligned history features of the three response and six condition channels,
and let $\overline{\bm C}_q\in\mathbb R^{N_\ell\times6}$ contain the target-time condition sampled at the same coordinates.
Its $\ell$th slot is
$\overline{\bm C}_{q,\ell}
=[\overline{\psi}_{q,\ell},
\overline{\Delta\psi}_{q,\ell},
\overline{\bm g}_{q,\ell}^{\top},
\overline{\bm u}_{\mathrm{bc},q,\ell}^{\top}]^{\top}$,
Here, overbars mark aligned values before standardization.
Vorticity determines the observed-frame partition but does not enter either tensor.
Fixed channelwise maps standardize both quantities after aggregation and alignment,
giving $\bm Z_r=\mathcal N_z(\overline{\bm Z}_r)$ and
$\bm C_q=\mathcal N_c(\overline{\bm C}_q)$.
The unstandardized nondimensional signed-distance component
$\overline{\bm\psi}_q\in\mathbb R^{N_\ell}$ is retained for boundary localization.
The query lead $\tau$ remains a separate scalar input.
Evaluating the prescribed field $\psi_q$ on $\bm X_q$ defines the query-frame fluid mask used for training and evaluation.
For backbone $b$, let $\mathcal E_b$, $\mathcal O_b$, and $\mathcal D_b$ denote its input adapter, core computation, and dense readout, and let
$\mathcal M_b^{\mathrm{in}}$ and $\mathcal M_b^{\mathrm{out}}$ denote its IN-DSFiLM and OUT-DSFiLM maps.
STCO computes
\begin{equation}
\begin{aligned}
    \bm h_0&=\mathcal E_b(\bm Z_r,\bm X_r,\tau),&
    \bm h_1&=\mathcal M_b^{\mathrm{in}}
      (\bm h_0,\bm C_q,\overline{\bm\psi}_q,\tau),\\
    \bm h_2&=\mathcal O_b(\bm h_1,\bm X_r,\tau),&
    \bm h_3&=\mathcal M_b^{\mathrm{out}}
      (\bm h_2,\bm C_q,\overline{\bm\psi}_q,\tau),\\
    \widehat{\bm y}_q(\bm X_q)
        &=\mathcal D_b(\bm h_3,\bm X_r,\bm X_q).
\end{aligned}
\label{eq:stco-pipeline}
\end{equation}
Here, $\bm h_0,\bm h_1\in\mathbb R^{N_\ell\times D_{\mathrm{in}}}$ and
$\bm h_2,\bm h_3\in\mathbb R^{N_\ell\times D_{\mathrm{out}}}$, where
$D_{\mathrm{in}}$ and $D_{\mathrm{out}}$ are the respective interface widths.
For the shared DSFiLM form,
$(\bm h_{\mathrm{in}},\bm h_{\mathrm{out}})=(\bm h_0,\bm h_1)$
for IN-DSFiLM and $(\bm h_2,\bm h_3)$ for OUT-DSFiLM.
The two modulation maps have independent parameters.

\paragraph{Flow-Aware Graph Leaf.}
FAGL maps each observed grid to $N_\ell$ adaptive regional slots through graph construction, partition, and remapping, as shown in Figure~\ref{fig:stco-overview}(b).
\textbf{Graph.}
At observed frame $n$, let
$\omega_n(\bm x)=\partial_xv_n(\bm x)-\partial_yu_n(\bm x)$
denote the scalar vorticity and $\mu_n(\bm x)=|\omega_n(\bm x)|$ its magnitude.
An eight-neighbor grid graph assigns edge dissimilarity from spatial separation and contrasts in $\mu_n$ and $\|\nabla\mu_n\|_2$.
Processing edges in increasing dissimilarity merges adjacent components when the connecting edge is small relative to their internal variation
\cite{felzenszwalb2004graphseg}.
This retains sharp vorticity changes for refinement.
\textbf{Partition.}
Each eligible component $C$ in refinement batch $\mathcal B$ is ranked by
\begin{equation}
\mathcal P_n(C)
    =\frac{|C|}{2}
     +\frac{\sum_{\bm x_i\in C}\mu_n(\bm x_i)}
            {\bar\mu_{\mathcal B}+\varepsilon_\mu}
\label{eq:fagl-priority}
\end{equation}
where $|C|$ is its number of grid points, $\bar\mu_{\mathcal B}$ is the mean of $\mu_n$ over $\mathcal B$, and $\varepsilon_\mu=10^{-6}$ stabilizes the denominator.
The two terms preserve spatial coverage and direct more slots to integrated vorticity activity.
Inspired by feature-driven refinement for vortex-dominated flows
\cite{kamkar2011feature},
selected components are median-split along the best of $x$, $y$, and signed $\omega_n$; the signed candidate separates opposing rotations when selected.
Refinement ends at $N_\ell$ or when no eligible split remains, after which the collection is normalized to exactly $N_\ell$ slots.
\textbf{Remap.}
For the grid points $\mathcal S_{n,\ell}$ in slot $\ell$, let $\mathcal A_{n,\ell}[\bm f]$ be the mean of field $\bm f$ and $\bm x_{n,\ell}$ their centroid.
Historical aggregates and the target-time condition are aligned to $\bm X_r$ by
\begin{equation}
\begin{aligned}
    \overline{\bm z}_{k,\ell}
        &=\operatorname{IDW}_4\!
          \left(\left\{
          \bm x_{n_k,j},
          \mathcal A_{n_k,j}[\bm y_{n_k},\bm c_{n_k}]
          \right\}_{j=1}^{N_\ell},
          \bm x_{r,\ell}\right),\\
    \overline{\bm C}_{q,\ell}
        &=\operatorname{IDW}_4[\bm c_q](\bm x_{r,\ell}).
\end{aligned}
\label{eq:fagl-regions}
\end{equation}
Here, $\operatorname{IDW}_4$ uses normalized inverse-distance weights over four nearest sources; the second line interpolates directly from the target-time condition mesh.
Stacking $\overline{\bm z}_{k,\ell}$ gives $\overline{\bm Z}_r$.
The supplementary material specifies the merge, refinement, interpolation, and slot-normalization rules.

\paragraph{Dual-Site Feature-wise Linear Modulation.}
Spatial alignment places prescribed mechanisms on common slots but leaves their influence on current features unresolved.
As detailed in Figure~\ref{fig:stco-overview}(c), DSFiLM combines feature-wise affine conditioning in FiLM
\cite{perez2018film}, spatially varying modulation in SPADE
\cite{park2019spade}, and feature-dependent channel gating in squeeze--excitation
\cite{hu2018senet}.
IN-DSFiLM acts before the backbone core and OUT-DSFiLM before dense readout.
The two sites have independent parameters.

Consider a slot at either site and omit site and slot indices.
Let $\bm h_{\mathrm{in}}\in\mathbb R^D$ be the feature entering that module and $\bm h_{\mathrm{out}}$ its modulated output.
Three independent two-layer networks encode standardized local
$\Delta\psi_q$, $\bm u_{\mathrm{bc},q}$, and $\bm g_q$ channels as the shifts
$\bm\beta_{\mathrm{mot}}$, $\bm\beta_{\mathrm{inflow}}$, and
$\bm\beta_{\mathrm{force}}\in\mathbb R^D$.
Motion acts near the body, so its shift is multiplied by
$w=\exp[-\overline{\psi}_q^{\,2}/(2\delta^2)]$.
$\overline{\psi}_q$ denotes the unstandardized signed distance at that slot and $\delta>0$ is a learned bandwidth.
A fourth two-layer network maps the sinusoidal encoding of $\tau$ to a lead-time scale
$\bm\gamma(\tau)\in\mathbb R^D$.
This scale is shared across slots.
A sigmoid affine projection of $\bm h_{\mathrm{in}}$ produces three channel-wise gates
$\bm\alpha_{\mathrm{mot}}$, $\bm\alpha_{\mathrm{inflow}}$, and
$\bm\alpha_{\mathrm{force}}\in(0,1)^D$.
The modulation is
\begin{equation}
\begin{aligned}
    \bm h_{\mathrm{out}}
    &=\bm h_{\mathrm{in}}\odot[\bm1+\bm\gamma(\tau)]
      +\bm\alpha_{\mathrm{mot}}\odot
       (w\bm\beta_{\mathrm{mot}})\\
    &\quad+\bm\alpha_{\mathrm{inflow}}\odot
       \bm\beta_{\mathrm{inflow}}
      +\bm\alpha_{\mathrm{force}}\odot
       \bm\beta_{\mathrm{force}},
\end{aligned}
\label{eq:dsfilm}
\end{equation}
where $\bm1\in\mathbb R^D$ is the all-ones vector and $\odot$ denotes elementwise multiplication.
Inspired by the identity-preserving zero-initialization principle of adaLN-Zero
\cite{peebles2023dit}, the final scale and shift layers are initialized at zero, so Equation~\ref{eq:dsfilm} begins with $\bm h_{\mathrm{out}}=\bm h_{\mathrm{in}}$.

\subsection{External-Activity Sampling}

Brief body-force and inflow events yield few condition-revealing pairs under uniform reference--query sampling.
For simulation $m$, let $a_m(n)\in[0,1]$ be the spatial mean of
$\|\bm g_{m,n}\|_2^2+\|\bm u_{\mathrm{bc},m,n}\|_2^2$ over grid points that are fluid in at least one stored frame, normalized by its maximum over that simulation.
For simulations without external force or inflow, $a_m(n)=0$ for every $n$.
Within the selected simulation, an admissible reference--query pair is sampled according to
\begin{equation}
    \Pr(n_r,\Delta n\mid m)
        \propto
        \eta_m(n_r,\Delta n)
        \left[\varepsilon+\sum_{j=0}^{\Delta n}a_m(n_r+j)\right].
\label{eq:transition-sampling}
\end{equation}
The sum accumulates external activity from the reference frame through the query frame.
The floor $\varepsilon=0.1$ gives every admissible pair nonzero probability.
The onset factor $\eta_m(n_r,\Delta n)=10$ when an interval crosses the first frame where any force or inflow component exceeds $10^{-3}$ in magnitude.
It is $1$ otherwise.
Motion remains part of $\bm c_q$ but is excluded from $a_m$, so the activity weighting targets force and inflow events.
Every matched Base--STCO pair receives identical temporal and spatial samples.
The supplementary material gives the complete rule.

\section{Experiments}
\label{sec:experiments}

\subsection{Experimental Setup}
\label{sec:experimental-setup}

\textbf{Datasets.}
The benchmark contains $142$ two-dimensional simulations of incompressible moving-body flow generated with \texttt{WaterLily.jl}
\cite{weymouth2025waterlily}.
Each simulation contains $256$ stored frames, giving $36{,}352$ frames in total.
All simulations use $\mathrm{Re}=5000$.
The benchmark spans five swimmer morphologies, an undisturbed baseline, and six prescribed-condition families.
Three families use analytic instantiations of Lamb--Oseen vortex encounters, convecting finite-width velocity gusts, and localized transverse gusts.
Their physical settings are motivated by prior experimental and computational studies
\cite{hufstedler2019vortical,bartels2013gustmodel,sedky2022gust}.
The remaining families comprise localized Gaussian body-force actuation, prescribed pose or undulation changes, and compound motion--disturbance cases.
The held-out simulations vary in amplitude, onset, placement, duration, waveform, and morphology.
The training, validation, and test sets contain $92$, $8$, and $42$ simulations, respectively, with no simulation shared across splits.
The reference swimming kinematics, analytic condition profiles, waveform variants, parameter values, and case allocation are given in the supplementary material.

\textbf{Backbones.}
We evaluate the same STCO interface on twelve architecture cores adapted to a common FAGL-to-query contract:
MGN~\cite{pfaff2021meshgraphnets},
RIGNO~\cite{mousavi2025rigno},
PINO~\cite{li2021pino},
Poseidon~\cite{herde2024poseidon},
GAOT~\cite{wen2025gaot},
GINO~\cite{li2023gino},
Transolver++~\cite{luo2025transolverpp},
Unisolver~\cite{zhou2024unisolver},
MPP~\cite{mccabe2024mpp},
CALM-PDE~\cite{hagnberger2025calmpde},
UPT~\cite{alkin2024upt}, and
GEPS~\cite{koupai2024geps}.
They span graph, spectral, geometry-aware, transformer, and latent operator architectures.
The labels denote adapted architecture cores rather than reproductions of the original training recipes.
All configurations are trained from scratch with the common data objective.
PINO uses its FNO core without the physics-informed loss, and Poseidon uses its ScOT core.
MPP denotes a Transformer core without multiple-physics pretraining, while GEPS uses low-rank context layers without inference-time adaptation.

\textbf{Implementation.}
Vorticity-aware FAGL represents each grid with $12{,}000$ adaptive slots, approximately $9.4\%$ of the mean fluid-grid cardinality.
Across five representative simulations, mean channelwise reconstruction errors range from $1.5\%$ to $11.8\%$.
With one observed frame, each epoch samples $10{,}000$ reference--query pairs with replacement at $\Delta n=1$--$20$.
Each pair resamples $15{,}000$ supervision coordinates, nominally comprising $80\%$ FAGL-region representatives and $20\%$ additional nonrepresentative locations.
After framewise outlier removal and fixed standardization, all models minimize the fluid-masked channel-mean relative-$L_2$ loss over standardized $u$, $v$, and $p$.
Training runs for $80$ epochs on one H200 GPU using AdamW with an initial learning rate of $5\times10^{-4}$ and batch size $32$.
In paired experiments, Base and STCO share the split, samples, FAGL, retained backbone context, adapters, decoder, query geometry, and optimization.
Base bypasses both DSFiLM sites, whereas STCO activates the complete interface.

\ResultsTableAndPerformanceFloat
\SpatialResultsFloat

\textbf{Evaluation.}
A predefined set of $5{,}004$ unique reference--query pairs from all $42$ test simulations is shared across configurations.
Its six $834$-pair strata combine three regimes (pre-event, motion transition, and external onset) with ID-Lead ($\Delta n=1$--$20$) or OOD-Lead ($\Delta n=21$--$40$).
All predictions are scored on target-frame fluid points of the complete $514\times514$ mesh.
Within each stratum, $E_y$ pools queries and fluid points per eligible simulation to compute channelwise relative-$L_2$ errors, then macro-averages over $u$, $v$, $p$, and simulations.
The ID-Lead, OOD-Lead, and overall scores average three, three, and six strata, respectively.
$E_{F_p}$ macro-averages per-simulation joint pressure-derived drag--lift RMSE normalized by fixed training-set RMS scales.
Both metrics use inverse-standardized predictions.
Latency is the batch-one H200 forward-pass time with dense decoding but without FAGL preprocessing.

Modulation counterfactual (MCF) sensitivity measures functional dependence on
$r\in\{\tau,\psi,\Delta\psi,\bm g,\bm u_{\mathrm{bc}}\}$.
It is evaluated on the same fixed set $\mathcal I$ of $5{,}004$ pairs shared by all model cells.
Let $\widehat{\bm Y}_{i,j}$ and $\widehat{\bm Y}^{(r)}_{i,j}$ denote the original and intervened inverse-standardized full-mesh predictions for channel $j$.
The reported score is
\begin{equation}
\operatorname{MCF}_r
=\frac{1}{3|\mathcal I|}
\sum_{i\in\mathcal I}\sum_{j\in\{u,v,p\}}
\frac{\left\|\bm M_i\odot
\left(\widehat{\bm Y}_{i,j}-\widehat{\bm Y}^{(r)}_{i,j}\right)\right\|_2}
{\max\!\left\{\left\|\bm M_i\odot\widehat{\bm Y}_{i,j}\right\|_2,10^{-6}\right\}} .
\label{eq:mcf}
\end{equation}
Here, $\bm M_i$ is the fixed fluid mask.
Each intervention holds the observed history, other condition groups, slot and query coordinates, fluid mask, and random state fixed.
MCF measures functional input dependence, complementing $E_y$ and $E_{F_p}$, which quantify accuracy against paired CFD ground truth.
The supplementary material specifies the load metric, complete aggregation, intervention distributions, and deterministic seeding.

\subsection{Main Results}

STCO improves accuracy for PCOL across twelve heterogeneous backbones (Table~\ref{tab:main-results}).
The matched gains evaluate the complete condition interface, rather than an individual route or modulation site.

STCO lowers $E_y$ in \STCOFieldWins{}/\STCOStrata{} regime--lead comparisons (mean: \STCOMeanFieldGain{}) and yields a positive six-stratum mean field gain for all \STCOAggregateWins{} backbones.
This includes all \STCOObservedContextWins{} backbones retaining observed-frame conditions and all \STCOTimeContextWins{} with existing lead-time inputs.
The complete interface therefore adds predictive value alongside the retained observed-condition and lead-time pathways.
Across the same comparisons, $E_{F_p}$ decreases by \STCOMeanLoadGain{} on average and improves in \STCOLoadWins{} cases.
The cross-backbone field gain is $40.4\%$ in ID-Lead and $23.1\%$ in OOD-Lead, while load reductions remain comparable at $19.3\%$ and $20.4\%$.

\paragraph{Resolved physical responses.}
Figure~\ref{fig:spatial-results} links the aggregate gains to resolved fields and pressure-derived loads.
On the common vortical query, STCO reduces eight of nine displayed channel errors, with lower residuals around the body, shear layers, and wake.
For prescribed yaw and transverse gust, the transverse-velocity error decreases by $62.8\%$ and $70.4\%$, respectively, at $\Delta n=20$.
Across the 40 direct queries in each case, lift and drag RMSE reductions span $33.6\%$--$75.1\%$.
Together, the motion and inflow examples connect improved condition-specific fields to pressure-derived responses relevant to downstream evaluation.

\paragraph{Longer leads, condition dependence, and cost.}
\STCOOODWins{} backbones retain lower OOD-Lead $E_y$ beyond the training range.
In Figure~\ref{fig:performance-summary}(b), the cross-backbone median remains below Base at all forty leads, with mean reductions of \STCOLagIDGain{} in ID-Lead and \STCOLagOODGain{} in OOD-Lead.
Although both medians rise overall across OOD-Lead, the advantage persists.
MCF is nonzero for every evaluated spatial condition group and STCO backbone, establishing condition dependence alongside the paired CFD accuracy metrics.
Spatial MCF peaks for GINO, while field gain peaks for GAOT, separating sensitivity from accuracy.
Figure~\ref{fig:performance-summary}(a) shows that these accuracy gains are achieved at comparable inference latency across matched pairs.

\paragraph{Limitations.}
The evaluation uses one seed and one observed frame in a two-dimensional moving-body Navier--Stokes system from one CFD solver.
$E_{F_p}$ excludes viscous shear, and the largest yaw excursions remain underestimated.
The load curves assemble independent target-time queries.
Complete condition trajectories, cross-time-consistent prediction, broader PDEs, and closed-loop action selection remain future work.

\section{Conclusion}

PCOL formulates future-response prediction from observed history and prescribed target-time fields.
STCO couples vorticity-aware FAGL, DSFiLM, and activity sampling across heterogeneous backbones.
Across twelve matched pairs, it reduces mean field and load error within and beyond the training lead range, with condition dependence at comparable inference cost.

\clearpage
\appendix
\section*{Supplementary Material}
\graphicspath{{figures/}{figures/supplementary/}}

This supplement details the benchmark, FAGL implementation, training, evaluation, and extended results.

\section{Benchmark}

Each simulation contains $T=256$ stored frames generated with the immersed-boundary computational fluid dynamics (CFD) solver \texttt{WaterLily.jl}~\cite{weymouth2025waterlily}.
The CFD response fields serve as ground truth (GT).
At frame $n$, the response and prescribed-condition fields are
$\bm y_n=[u_n,v_n,p_n]^\top$ and
$\bm c_n=[\psi_n,\Delta\psi_n,\bm g_n^\top,\bm u_{\mathrm{bc},n}^\top]^\top$.
The reference and query indices satisfy $n_q=n_r+\Delta n$, with normalized lead $\tau=\Delta n/(T-1)$.
All experiments use the one-frame history $\mathcal H_r=\{(\bm y_r,\bm c_r)\}$.

\subsection{Simulation Families and Splits}

\begin{figure*}[t]
    \centering
    \includegraphics[width=0.96\textwidth]{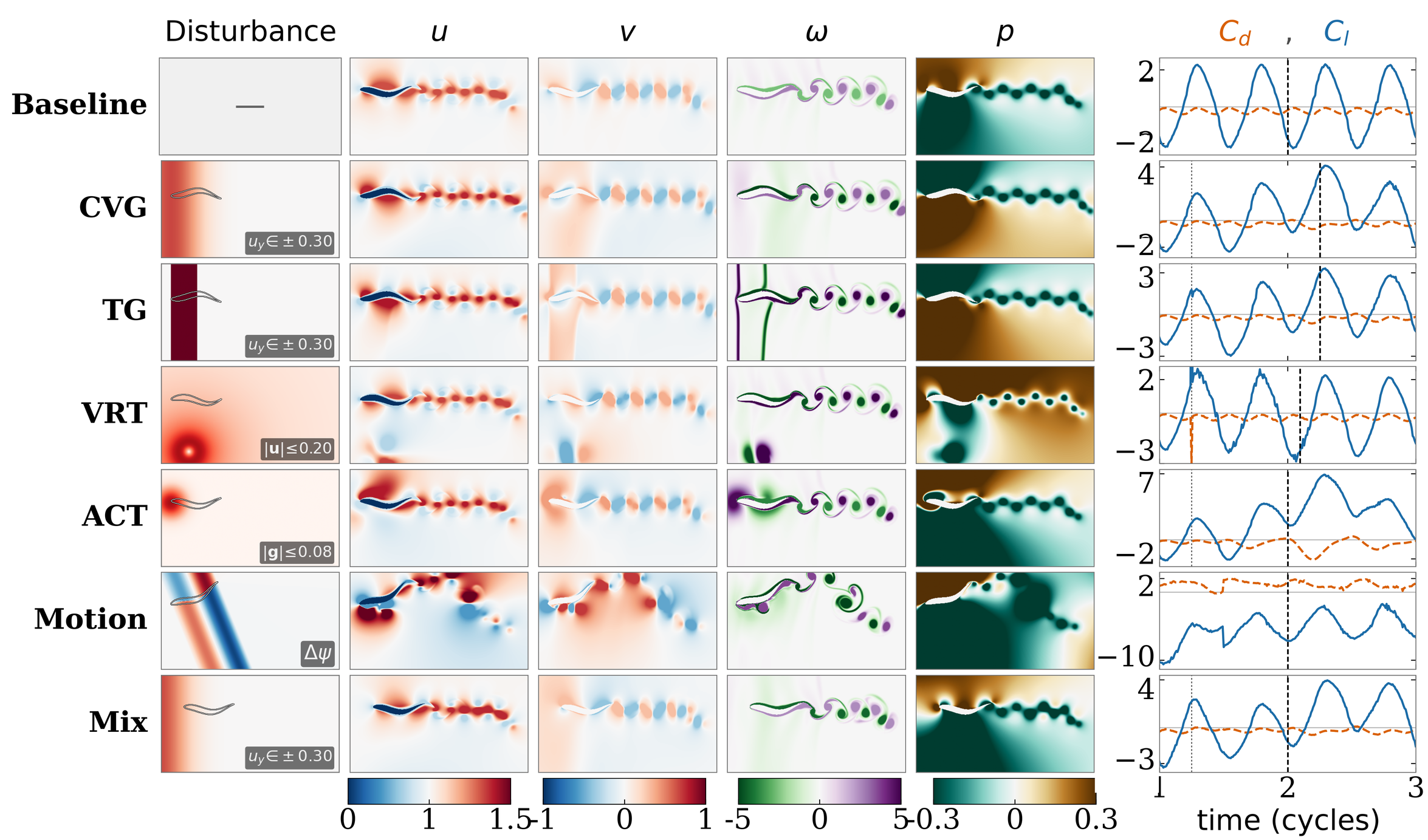}
    \caption{Representative benchmark simulations. Rows correspond to the undisturbed, convecting-gust, transverse-gust, vortical-encounter, actuation, motion, and compound families. The first column shows prescribed condition fields, with motion represented by the one-frame signed-distance change $\Delta\psi$. The remaining columns show CFD GT fields and CFD load histories. Dashed markers identify the displayed frames. The learned target is the fluid-region velocity--pressure field $(u,v,p)$.}
    \label{fig:supp-dataset-gallery}
\end{figure*}

Figure~\ref{fig:supp-dataset-gallery} illustrates the undisturbed baseline and six prescribed-condition families through representative condition fields, CFD responses, and load histories.
Table~\ref{tab:supp-family-composition} gives the $92/8/42$ simulation-level train/validation/test allocation.

\subsection{Flow, Kinematics, and Prescribed Conditions}

All simulations use the same base inflow, $\mathrm{Re}=5000$, NACA~0012 section, and solver configuration.
The $514\times514$ grid resolves one chord with $L_{\mathrm{pix}}=128$ lattice units, and the nondimensional variables set $L=U_\infty=\rho=1$.
For body-frame chordwise coordinate $x$, define $s=\operatorname{clip}(x/L,0,1)$.
The reference centerline is
\begin{equation}
\begin{aligned}
    y_b(s,t)
        &=L A_b(s)
          \cos\!\left(\frac{2\pi s}{\lambda^+}-\omega_{\mathrm{kin}}t+\phi\right),\\
    A_b(s)
        &=A_{0,b}+A_{1,b}s+A_{2,b}s^2,\\
    \omega_{\mathrm{kin}}
        &=\frac{2\pi \mathrm{St}\,U_\infty}{2A_{\max}L}.
\end{aligned}
\label{eq:supp-fish-kinematics}
\end{equation}
Here, $b$ indexes morphology, $A_b(s)$ is its dimensionless amplitude envelope, $\lambda^+$ is the nondimensional body wavelength, and $\phi$ is the phase.
Table~\ref{tab:supp-morphology-coefficients} lists five amplitude envelopes with tail amplitude $A_{\max}=A_b(1)=0.1$.
All cases use Strouhal number $\mathrm{St}=0.4$, $\lambda^+=1$, and $\phi=0$.
The kinematic period is $T_{\mathrm{cyc}}=2\pi/\omega_{\mathrm{kin}}$.

\begin{samepage}
\begin{equation}
    v_{\mathrm{gust}}(x,t)
        =A_gF(\xi),\qquad
          \xi=\frac{U_\infty(t-t_0)-(x-x_0)}{w}.
\label{eq:supp-gust-profile}
\end{equation}
Here, $v_{\mathrm{gust}}$ is the prescribed transverse-velocity perturbation, $t_0$ is the onset time, $x_0$ and $w$ are the streamwise reference position and profile scale, and $A_g$ is the peak amplitude.
\end{samepage}
Convecting gusts use Gaussian, sine, or one-minus-cosine profiles, and transverse gusts use top-hat, sine-squared, or trapezoidal profiles
\cite{bartels2013gustmodel,andreuangulo2020gustprofiles,sedky2022gustcomparison}; the Gaussian profile is $F(\xi)=e^{-4\ln2(\xi-1/2)^2}$.

The Lamb--Oseen vortical inflow is
\begin{equation}
    \bm u_{\mathrm{vrt}}(\bm x,t)
        =\frac{\Gamma}{2\pi r^2}
          \left(1-e^{-\alpha r^2/r_c^2}\right)\bm r^\perp.
\label{eq:supp-vortex-profile}
\end{equation}
Here, $\bm x=(x,y)$, $\Gamma$ is the circulation, $r_c$ is the core radius, and $(x_0,y_0)$ is the vortex center at $t=t_0$.
With $\bm x_c(t)=(x_0+U_\infty(t-t_0),y_0)$, define $\bm r=\bm x-\bm x_c(t)$, $r=\|\bm r\|_2$, and $\bm r^\perp=(-r_y,r_x)$.
The vortex is zero for $t<t_0$, and $\alpha=1.25643$ places the peak tangential velocity at $r=r_c$.

Body-force actuation uses
\begin{equation}
    \bm g(\bm x,t)
        =A_f(t)
          e^{-\|\bm x-\bm x_f\|_2^2/(2\sigma^2)}\bm e_y.
\label{eq:supp-actuation-profile}
\end{equation}
Here, $\bm x_f$ and $\sigma$ are the forcing center and spatial scale, and $\bm e_y$ is the transverse unit vector.
The one-cosine amplitude is $A_f(t)=\tfrac12A[1-\cos(2\pi(t-t_0)/d)]$ on $[t_0,t_0+d]$ and zero otherwise, where $A$ and $d$ are its peak amplitude and duration.

Prescribed motion uses
\begin{equation}
    \chi(t)
        =\chi_\infty\mathcal R\!\left(\frac{t-t_0}{d}\right).
\label{eq:supp-motion-profile}
\end{equation}
Here, $\chi(t)$ is the displacement or yaw angle, $\chi_\infty$ is its terminal value, and $d$ is the ramp duration.
$\mathcal R(\vartheta)$ equals $0$ for $\vartheta\leq0$, $\tfrac12(1-\cos\pi\vartheta)$ for $0<\vartheta<1$, and $1$ for $\vartheta\geq1$.
At target frame $q$, the gust and vortex fields define $\bm u_{\mathrm{bc},q}$, actuation defines $\bm g_q$, and prescribed motion determines $\psi_q$ and $\Delta\psi_q=\psi_q-\psi_{q-1}$.
Table~\ref{tab:supp-condition-parameters} summarizes the varied factors and profiles.

\begin{table}[t]
    \centering
    {\small
    \setlength{\tabcolsep}{1.6pt}
    \begin{tabular*}{\columnwidth}{@{\extracolsep{\fill}}llrrrr@{}}
        \toprule
        Label & Condition & Train & Val. & Test & Total \\
        \midrule
        \texttt{baseline} & Undisturbed & 4 & 1 & 0 & 5 \\
        \texttt{vrt} & Vortical encounter & 15 & 2 & 7 & 24 \\
        \texttt{cvg} & Convecting gust & 16 & 1 & 9 & 26 \\
        \texttt{tg} & Transverse gust & 17 & 1 & 8 & 26 \\
        \texttt{act} & Body-force actuation & 18 & 1 & 10 & 29 \\
        \texttt{motion} & Prescribed motion & 19 & 1 & 4 & 24 \\
        \texttt{mix} & Motion + disturbance & 3 & 1 & 4 & 8 \\
        \midrule
        Total & & 92 & 8 & 42 & 142 \\
        \bottomrule
    \end{tabular*}
    }
    \caption{Composition of the moving-body-flow benchmark. Training, validation, and test contain $92$, $8$, and $42$ simulations. The test set covers all six prescribed-condition families.}
    \label{tab:supp-family-composition}
\end{table}

\begin{table}[t]
    \centering
    {\small
    \setlength{\tabcolsep}{3pt}
    \begin{tabular*}{\columnwidth}{@{\extracolsep{\fill}}lrrr@{}}
        \toprule
        Morphology $b$ & $A_{0,b}$ & $A_{1,b}$ & $A_{2,b}$ \\
        \midrule
        Anguilliform & 0.0367 & 0.0323 & 0.0310 \\
        Carangiform & 0.0200 & $-0.0825$ & 0.1625 \\
        Subcarangiform & 0.0200 & 0.0300 & 0.0500 \\
        Thunniform & 0.0050 & $-0.0200$ & 0.1150 \\
        Ostraciiform & 0.0020 & $-0.0350$ & 0.1330 \\
        \bottomrule
    \end{tabular*}
    }
    \caption{Centerline-amplitude coefficients in Equation~\ref{eq:supp-fish-kinematics}. All five profiles satisfy $A_b(1)=A_{\max}=0.1$ while varying the chordwise amplitude envelope.}
    \label{tab:supp-morphology-coefficients}
\end{table}

\begin{table*}[t]
    \centering
    {\small
    \setlength{\tabcolsep}{1.5pt}
    \begin{tabular*}{0.95\textwidth}{@{\extracolsep{\fill}}lp{0.54\textwidth}p{0.28\textwidth}@{}}
        \toprule
        Family & Varied benchmark factors & Profile or command \\
        \midrule
        \texttt{vrt}
        & Circulation, lateral placement, and onset
        & Lamb--Oseen \\
        \texttt{cvg}
        & Amplitude, streamwise placement, and onset
        & Gaussian, sine, one-minus-cosine \\
        \texttt{tg}
        & Amplitude, streamwise placement, and onset
        & Top-hat, sine-squared, trapezoid \\
        \texttt{act}
        & Amplitude, placement, onset, and duration
        & One-cosine, step, ramp-and-hold \\
        \texttt{motion}
        & Surge, sway, and yaw magnitude
        & Smooth ramp \\
        \texttt{mix}
        & Motion--disturbance pairing and relative onset
        & \texttt{cvg}$+$surge, \texttt{act}$+$surge, \texttt{vrt}$+$yaw, \texttt{tg}$+$sway \\
        \bottomrule
    \end{tabular*}
    }
\caption{Prescribed-condition families and varied benchmark factors. Compound cases pair one motion command with one disturbance.}
    \label{tab:supp-condition-parameters}
\end{table*}

\subsection{Frames, Candidate Pairs, and Preprocessing}

With $256$ frames per simulation, the benchmark contains $142\times256=36{,}352$ frames, including $92\times256=23{,}552$ for training.
The $216$ reference indices $n_r=20,\ldots,235$ and $20$ training leads $\Delta n=1,\ldots,20$ give $92\times216\times20=397{,}440$ admissible training pairs.
Each epoch samples $10{,}000$ pairs with replacement.

After framewise response outlier removal, fixed channelwise maps standardize the aligned response and prescribed-condition inputs.
Metrics use inverse-standardized response fields.

\section{FAGL Implementation}

FAGL derives the observed vorticity
$\omega_r=\partial_xv_r-\partial_yu_r$ and its magnitude
$\mu_r=|\omega_r|$ from $\bm y_r$.
For simulation $m$, let $\mathcal V_m$ denote the time-union fluid grid determined by its prescribed geometry sequence.
The observed-frame regional slots are
\begin{equation}
    \{\mathcal S_{r,\ell}\}_{\ell=1}^{N_\ell}
    =\Pi_{N_\ell}(\mu_r,\mathcal V_m),
    \qquad N_\ell=12{,}000,
\label{eq:supp-observed-partition}
\end{equation}
where $\Pi_{N_\ell}$ denotes the FAGL partitioner and $\mathcal S_{r,\ell}$ contains the grid points assigned to slot $\ell$.
Observed response and condition fields are averaged within each slot.
The prescribed target-time fields $(\psi_q,\Delta\psi_q,\bm g_q,\bm u_{\mathrm{bc},q})$ are aligned to the slot centroids by four-neighbor inverse-distance interpolation.

FAGL combines graph merging with bounded refinement to form a fixed-cardinality representation.
On the eight-neighbor grid graph, an edge $(i,j)$ receives dissimilarity $d_{ij}$ from grid distance, the contrast in observed vorticity magnitude $\mu_r$, and the contrast in $\|\nabla\mu_r\|_2$.
The three cues are normalized within each frame.

Edges are scanned once in ascending $d_{ij}$.
Components $A$ and $B$ merge when
\begin{equation}
d_{ij}\leq
\min\!\left\{
\operatorname{Int}(A)+\frac{\kappa}{|A|},
\operatorname{Int}(B)+\frac{\kappa}{|B|}
\right\},
\label{eq:supp-fagl-merge}
\end{equation}
where $\operatorname{Int}(C)$ is the largest accepted edge within $C$ and
$\kappa$ is the frame-adaptive merge threshold.

If the merge returns fewer than $N_\ell$ regions, refinement prioritizes
\begin{equation}
\mathcal P_r(C)=\frac{|C|}{2}
+\frac{\sum_{\bm x_i\in C}\mu_r(\bm x_i)}
{\bar\mu_{\mathcal B}+10^{-6}}.
\label{eq:supp-fagl-priority}
\end{equation}
Here, $|C|$ is the number of points in region $C$, and $\bar\mu_{\mathcal B}$ is the pointwise mean of $\mu_r$ over refinement batch $\mathcal B$.
Regions with at least four points are median-split along the best of $x$, $y$, and signed $\omega_r$, selected by balance and vorticity contrast.
Refinement stops at $N_\ell$ regions or when no eligible split remains, after which the collection is normalized to exactly $N_\ell$ slots.
Counts above $N_\ell$ are reduced by retaining the largest regions; counts below $N_\ell$ are padded with empty observed slots centered on the grid before target conditions are aligned.

Each nonempty slot stores the mean coordinate and observed fields of its member points.

\begin{figure*}[t]
    \centering
    \includegraphics[width=0.96\textwidth]{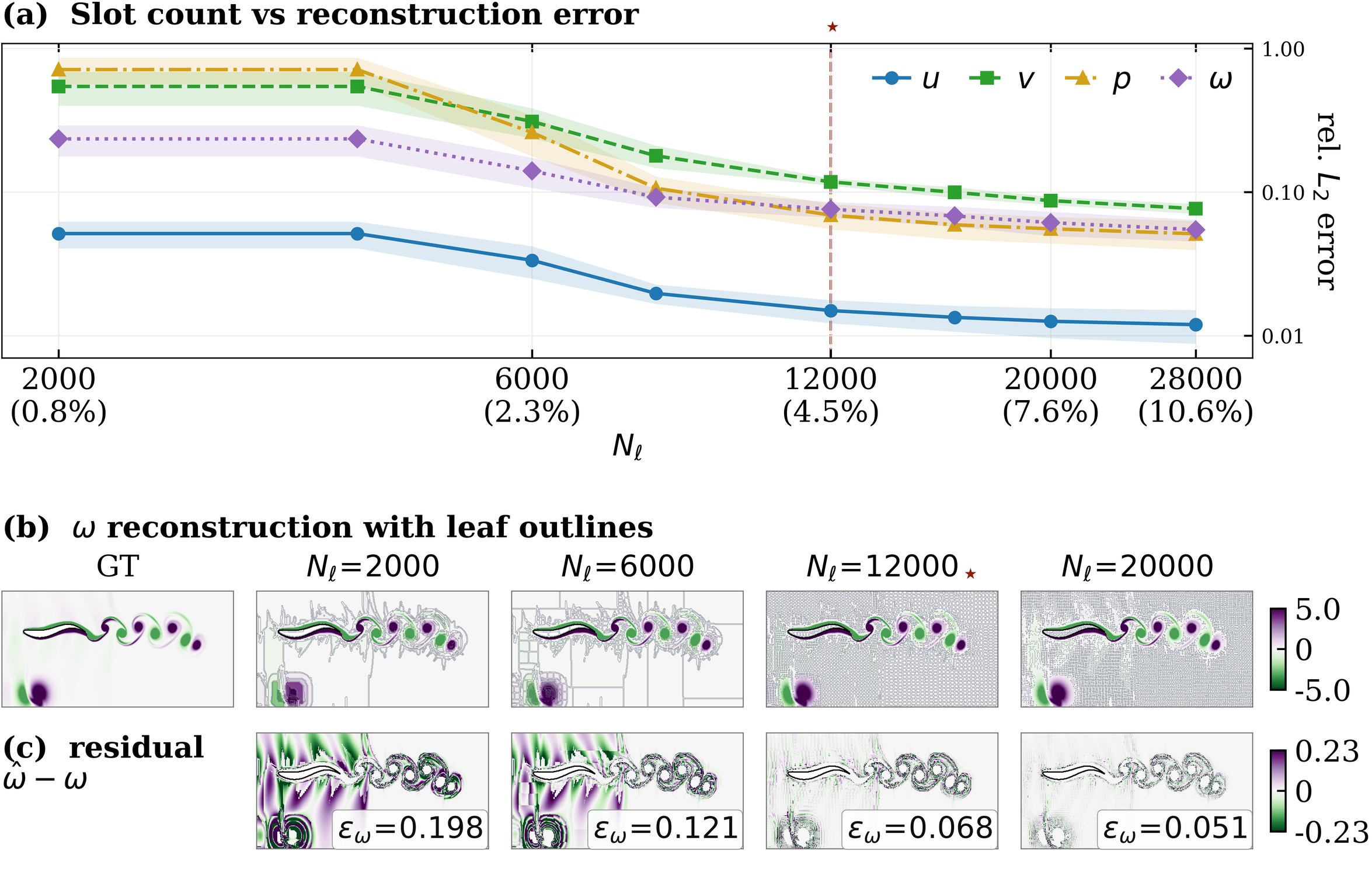}
    \caption{FAGL reconstruction across slot counts. In panel (a), parenthetical labels give $N_\ell$ as a percentage of the complete $514\times514$ mesh. (a) Fluid-region relative-$L_2$ error for constant-region reconstructions. Curves and bands give the mean and one standard deviation over five simulations at frame 128. At $N_\ell=12{,}000$, the mean errors are $1.5\%$, $11.8\%$, $6.9\%$, and $7.6\%$ for $u$, $v$, $p$, and $\omega_r$. (b) GT vorticity and representative regional reconstructions with region boundaries. (c) Signed residuals $\widehat\omega_r-\omega_r$. FAGL uses $\omega_r$ for partitioning, while the operator predicts $(u,v,p)$.}
    \label{fig:supp-fagl-reconstruction}
\end{figure*}

All interpolation and dense decoding use four-neighbor inverse-distance weighting,
\begin{equation}
\begin{aligned}
\operatorname{IDW}_4[\bm f](\bm x)
&=\sum_{j\in\mathcal N_4(\bm x)}w_j(\bm x)\bm f_j,\\
w_j(\bm x)
&=\frac{\rho_j(\bm x)^{-1}}
{\sum_{k\in\mathcal N_4(\bm x)}\rho_k(\bm x)^{-1}},\\
\rho_j(\bm x)&=\max\{\|\bm x-\bm x_j\|_2,10^{-8}\}.
\end{aligned}
\label{eq:supp-idw}
\end{equation}
Here, $\mathcal N_4(\bm x)$ indexes the four nearest source coordinates and $\bm f_j$ is the source value at $\bm x_j$.
FAGL uses $\omega_r$ only to construct the observed-frame partition.
The learned operator receives regional means of
$(u,v,p,\psi,\Delta\psi,\bm g,\bm u_{\mathrm{bc}})$.
Figure~\ref{fig:supp-fagl-reconstruction} reports reconstruction across slot counts.

\section{Training and Evaluation}

To isolate the predictive contribution of the target-time condition interface, each matched Base--STCO pair uses the same split, temporal and spatial samples, FAGL representation, retained backbone context, adapters, decoder, query geometry, and optimization.
Base bypasses both DSFiLM sites, whereas STCO activates the complete interface with $\bm c_q$ and $\tau$.

\subsection{External-Activity Sampling and Optimization}

External force and inflow disturbances are active over only part of each trajectory, so uniform pair sampling would draw many intervals without external activity.
We therefore sample each training example in two stages: first a simulation, and then a reference--query pair within that simulation.

For simulation $m$, let $\mathcal V_m$ denote its time-union fluid grid, $|\mathcal V_m|$ its number of points, and $h$ a grid-point index.
The fields $\bm g_{m,n,h}$ and $\bm u_{\mathrm{bc},m,n,h}$ are the prescribed body-force acceleration and inflow perturbation at frame $n$ and point $h$.
We define the case-normalized external activity at frame $n$ as
\begin{equation}
\begin{aligned}
\widetilde a_m(n)
&=\frac{1}{|\mathcal V_m|}\sum_{h\in\mathcal V_m}
\left(\|\bm g_{m,n,h}\|_2^2+
\|\bm u_{\mathrm{bc},m,n,h}\|_2^2\right),\\
a_m^\star&=\max_{n'}\widetilde a_m(n'),\\
a_m(n)&=
\begin{cases}
\widetilde a_m(n)/a_m^\star,&a_m^\star>10^{-12},\\
0,&\text{otherwise}.
\end{cases}
\end{aligned}
\label{eq:supp-activity-trace}
\end{equation}
Here, $\widetilde a_m(n)$ is the spatially averaged raw activity, $a_m^\star$ is its maximum over frames $n'$, and $a_m(n)\in[0,1]$ is the normalized activity trace.
The $10^{-12}$ threshold avoids division by a numerically zero maximum.
Prescribed body motion is excluded from this sampling score and remains represented by $\psi$ and $\Delta\psi$ in the model input.

At the first stage, undisturbed and zero-external-activity simulations receive weight $b_m=0.3$, while all others receive $b_m=1$.
We sample $m$ with
$\Pr(m)=b_m/\sum_{m'\in\mathcal M_{\mathrm{train}}}b_{m'}$,
where $\mathcal M_{\mathrm{train}}$ is the training simulation set and $m'$ is its summation index.

\begin{figure*}[t]
    \centering
    \includegraphics[width=0.95\textwidth]{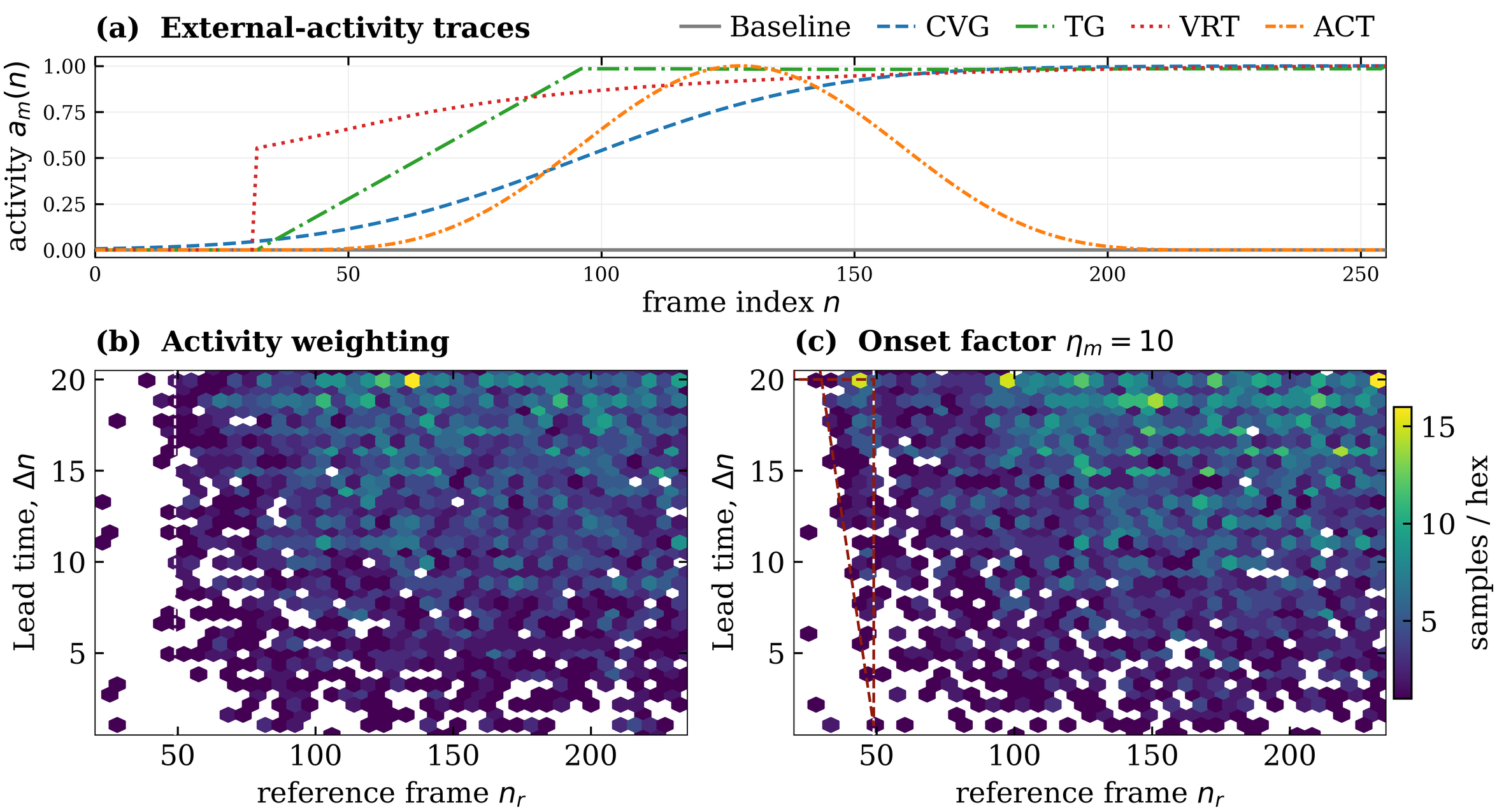}
    \caption{External-activity sampling. (a) Case-normalized traces for the undisturbed baseline, convecting gust (CVG), transverse gust (TG), Lamb--Oseen vortex (VRT), and body-force actuation (ACT). (b,c) Four thousand pairs sampled from one transverse-gust simulation. Activity weighting favors active intervals, and the onset factor further favors onset-crossing pairs.}
    \label{fig:supp-activity-sampling}
\end{figure*}

At the second stage, the admissible pairs are
$\mathcal A_m=\{(n_r,\Delta n):n_r\in\{20,\ldots,235\},\ \Delta n\in\{1,\ldots,20\}\}$.
$n_r$ is the reference frame and $\Delta n$ is the forecast lead in stored frames.
If simulation $m$ contains an external-activity onset, let $n_{\mathrm{on},m}$ be the first frame at which the absolute value of any component of $\bm g$ or $\bm u_{\mathrm{bc}}$ exceeds $10^{-3}$ anywhere on the grid, and define
\begin{equation}
\begin{aligned}
\eta_m(n_r,\Delta n)
&=
\begin{cases}
10,&n_r<n_{\mathrm{on},m}\leq n_r+\Delta n,\\
1,&\text{otherwise},
\end{cases}\\
w_m(n_r,\Delta n)
&=\eta_m(n_r,\Delta n)
\left[0.1+\sum_{j=0}^{\Delta n}a_m(n_r+j)\right],\\
\Pr(n_r,\Delta n\mid m)
&=\frac{w_m(n_r,\Delta n)}
{\sum_{(n'_r,\Delta n')\in\mathcal A_m}w_m(n'_r,\Delta n')}.
\end{aligned}
\label{eq:supp-activity-probability}
\end{equation}
Here, $\eta_m$ is the onset-crossing multiplier, $w_m$ is the unnormalized pair weight, $j$ indexes frames after $n_r$, and $(n'_r,\Delta n')$ indexes candidate pairs in the normalizing sum.
For simulations without an external-activity onset, $\eta_m=1$ for every pair.
The activity sum covers the complete reference-to-query interval; the constant $0.1$ retains nonzero probability for quiescent intervals, and $\eta_m$ increases the probability of intervals that cross the first onset.
The complete sampling distribution is
$\Pr(m,n_r,\Delta n)=\Pr(m)\Pr(n_r,\Delta n\mid m)$.
Figure~\ref{fig:supp-activity-sampling} illustrates the resulting activity and onset weighting.

Each sampled reference--query pair draws $15{,}000$ supervised coordinates.
The sampler targets $80\%$ unique representatives of observed-frame FAGL regions.
The remaining nominal budget comprises $12\%$ nonrepresentative locations weighted by observed $|\omega_r|$ and $8\%$ sampled uniformly.
All evaluation metrics use the full mesh.

Training uses batch size $32$, $80$ epochs, and AdamW with learning rate $5\times10^{-4}$, weight decay $10^{-3}$, $(\beta_1,\beta_2)=(0.9,0.95)$, gradient clipping at $1$, and cosine decay to $10^{-5}$.
Let $\widetilde{\bm y}_j,\widehat{\widetilde{\bm y}}_j\in\mathbb R^{B\times N_q}$ collect the standardized target and prediction for channel $j$ over a batch, and let $\bm M\in\{0,1\}^{B\times N_q}$ be the corresponding query-frame fluid mask.
Here, $B$ is the batch size, $N_q$ is the number of supervised coordinates, and $\theta$ denotes the trainable parameters.
The loss is
\begin{equation}
\mathcal L(\theta)
=\frac{1}{3}\sum_{j\in\{u,v,p\}}
\frac{\|\bm M\odot(\widehat{\widetilde{\bm y}}_j-\widetilde{\bm y}_j)\|_2}
{\|\bm M\odot\widetilde{\bm y}_j\|_2}.
\label{eq:supp-training-loss}
\end{equation}
With incomplete batches dropped, the $10{,}000$ training pairs give $312$ optimization steps per epoch.
Each matched Base--STCO pair uses the same training and validation pair-sampling seeds.
Every five epochs, validation evaluates $1{,}000$ sampled pairs from the eight validation simulations at $\Delta n=1$--$20$, using $15{,}000$ coordinates per pair.
Each result uses the checkpoint with the lowest validation relative-$L_2$ error.
Each of the 24 configurations is trained from scratch on a single NVIDIA H200 GPU.

\subsection{Predefined Evaluation Set}

Evaluation scores all target-frame fluid points on the complete $514\times514$ mesh.
The predefined evaluation set contains $5{,}004$ unique reference--query pairs from all $42$ test simulations, divided into six $834$-pair regime--lead strata and shared across all configurations.
ID-Lead is $\Delta n=1$--$20$, and OOD-Lead is $\Delta n=21$--$40$.
Pairs are assigned to the pre-event regime when $n_r+\Delta n<n_{\mathrm{event}}$, to motion transition when $n_{\mathrm{start}}\leq n_r<n_r+\Delta n\leq n_{\mathrm{end}}$, and to external onset when $n_r<n_{\mathrm{event}}\leq n_r+\Delta n$.
Here, $n_{\mathrm{event}}$ is the annotated disturbance onset, while $n_{\mathrm{start}}$ and $n_{\mathrm{end}}$ delimit the prescribed motion transition.
Table~\ref{tab:supp-test-manifest} gives the stratum composition.

\begin{table}[t]
    \centering
    {\small
    \setlength{\tabcolsep}{2.2pt}
    \begin{tabular*}{\columnwidth}{@{\extracolsep{\fill}}lrr@{}}
        \toprule
        Regime & \shortstack{ID-Lead\\pairs/cases}
        & \shortstack{OOD-Lead\\pairs/cases} \\
        \midrule
        Pre-event & $834/38$ & $834/8$ \\
        Motion transition & $834/4$ & $834/4$ \\
        External onset & $834/37$ & $834/37$ \\
        \bottomrule
    \end{tabular*}
    }
    \caption{Composition of the predefined $5{,}004$-pair evaluation set. Each regime--lead stratum contains $834$ pairs. Metrics macro-average the contributing simulations within each stratum. A simulation may contribute pairs to more than one regime.}
    \label{tab:supp-test-manifest}
\end{table}

\subsection{Field and Pressure-Derived Load Metrics}

Let $\varsigma$ index a regime--lead stratum, $\mathcal C_\varsigma$ its contributing simulations, and $\mathcal Q_{m,\varsigma}$ the selected pair indices from simulation $m$.
For channel $j\in\{u,v,p\}$ and mesh point $h$, let $y_{m,i,h,j}$ and $\widehat y_{m,i,h,j}$ be the GT and predicted target values, and let $M_{m,i,h}$ be the query-frame fluid mask.
For values $z_{m,i,h}$ on these pairs, define the masked norm
$\|z\|_{m,\varsigma}^{2}
=\sum_{i\in\mathcal Q_{m,\varsigma}}\sum_h
M_{m,i,h}|z_{m,i,h}|^2$.
The field metric is
\begin{equation}
\begin{aligned}
E_{m,\varsigma,j}
&=\frac{\|\widehat y_{m,j}-y_{m,j}\|_{m,\varsigma}}
{\max\{\|y_{m,j}\|_{m,\varsigma},10^{-6}\}},\\
E_y^{(\varsigma)}
&=\frac{1}{3|\mathcal C_\varsigma|}
\sum_{m\in\mathcal C_\varsigma}
\sum_{j\in\{u,v,p\}}E_{m,\varsigma,j}.
\end{aligned}
\label{eq:supp-field-metric}
\end{equation}

The pressure-derived drag and lift coefficients are reconstructed as
\[
\begin{aligned}
\bm C_p(t)
&=\begin{bmatrix}C_{D,p}(t)&C_{L,p}(t)\end{bmatrix}^{\top}\\
&=-\int_{\partial\Omega_b(t)}
p(\bm x,t)\bm n(\bm x,t)\,\mathrm ds .
\end{aligned}
\]
Here, $\partial\Omega_b(t)$ is the prescribed body boundary, $\bm n$ its outward unit normal, and $\mathrm ds$ the boundary arc-length element.
GT and predicted loads use the same query-frame signed-distance contour and quadrature.
The stored field is the pressure coefficient $p$ defined in the main paper.
For $k\in\{D,L\}$, the normalized pressure-derived load error is
\begin{equation}
\begin{aligned}
E_{m,\varsigma,F_p}^{\,2}
&=\frac{1}{2|\mathcal Q_{m,\varsigma}|}
\sum_{i\in\mathcal Q_{m,\varsigma}}\sum_{k\in\{D,L\}}\\
&\quad
\left(\frac{\widehat C_{m,i,k,p}-C_{m,i,k,p}}{s_k}\right)^2,\\
E_{F_p}^{(\varsigma)}
&=\frac{1}{|\mathcal C_\varsigma|}
\sum_{m\in\mathcal C_\varsigma}E_{m,\varsigma,F_p},
\end{aligned}
\label{eq:supp-force-metric}
\end{equation}
where $C_{m,i,k,p}$ is pressure-derived component $k$ for pair $i$ from simulation $m$, and $s_D=0.1187$ and $s_L=0.7959$ are RMS scales over all $397{,}440$ admissible training pairs.

For either $E\in\{E_y,E_{F_p}\}$, band and overall errors are
\begin{equation}
\begin{aligned}
E^{\mathrm{ID}}
&=\frac13\sum_{\varsigma\in\mathcal S_{\mathrm{ID}}}E^{(\varsigma)},&
E^{\mathrm{OOD}}
&=\frac13\sum_{\varsigma\in\mathcal S_{\mathrm{OOD}}}E^{(\varsigma)},\\
E^{\mathrm{all}}
&=\frac16\sum_{\varsigma}E^{(\varsigma)}.&&
\end{aligned}
\label{eq:supp-regime-aggregation}
\end{equation}
Here, $\mathcal S_{\mathrm{ID}}$ and $\mathcal S_{\mathrm{OOD}}$ contain the three regime strata in their respective lead bands.
The aggregation gives equal weight to contributing simulations and regime strata, while $E_y$ also weights the three response channels equally.
Paired gain is
$100(E_{\mathrm{Base}}-E_{\mathrm{STCO}})/E_{\mathrm{Base}}$.

\begin{table*}[t]
\section{Extended Results}
\centering
{\small
\setlength{\tabcolsep}{1.65pt}
\begin{tabular*}{0.95\textwidth}{@{\extracolsep{\fill}}llrrrrrrrrrrr@{}}
\toprule
& & \multicolumn{3}{c}{ID-Lead gain (\%)} & \multicolumn{3}{c}{OOD-Lead gain (\%)} & \multicolumn{5}{c}{Route-level MCF} \\
\cmidrule(lr){3-5}\cmidrule(lr){6-8}\cmidrule(l){9-13}
Backbone & Metric & Pre-event & Motion & External & Pre-event & Motion & External & $\tau$ & $\psi$ & $\Delta\psi$ & $\bm g$ & $\bm u_{\mathrm{bc}}$ \\
\midrule
MGN & Field & 63.9 & 24.4 & 43.0 & 21.2 & \textbf{$-7.4$} & 15.9 & \multirow{2}{*}{0.386} & \multirow{2}{*}{0.239} & \multirow{2}{*}{0.156} & \multirow{2}{*}{0.172} & \multirow{2}{*}{0.510} \\
    & Load  & 77.3 & 8.4 & 9.6 & 48.2 & 3.9 & 30.5 & & & & & \\
\addlinespace[1pt]
RIGNO & Field & 11.2 & 8.2 & 19.1 & 8.8 & 3.8 & 17.3 & \multirow{2}{*}{0.318} & \multirow{2}{*}{0.343} & \multirow{2}{*}{0.178} & \multirow{2}{*}{0.270} & \multirow{2}{*}{0.526} \\
      & Load  & 11.3 & \textbf{$-3.3$} & 2.2 & \textbf{$-11.6$} & \textbf{$-8.8$} & \textbf{$-2.5$} & & & & & \\
\addlinespace[1pt]
PINO & Field & 70.1 & 39.1 & 54.0 & 61.2 & 39.1 & 58.1 & \multirow{2}{*}{0.211} & \multirow{2}{*}{0.339} & \multirow{2}{*}{0.186} & \multirow{2}{*}{0.401} & \multirow{2}{*}{0.458} \\
     & Load  & 83.7 & 33.5 & 18.7 & 55.9 & 33.1 & 52.7 & & & & & \\
\addlinespace[1pt]
Poseidon & Field & 35.2 & 6.4 & 36.7 & 21.0 & 4.4 & 41.3 & \multirow{2}{*}{0.293} & \multirow{2}{*}{0.255} & \multirow{2}{*}{0.145} & \multirow{2}{*}{0.383} & \multirow{2}{*}{0.385} \\
         & Load  & 42.2 & \textbf{$-0.2$} & 5.2 & 18.9 & 2.4 & 34.8 & & & & & \\
\addlinespace[1pt]
GAOT & Field & 83.2 & 51.5 & 63.5 & 65.3 & 44.5 & 61.3 & \multirow{2}{*}{0.232} & \multirow{2}{*}{0.225} & \multirow{2}{*}{0.209} & \multirow{2}{*}{0.505} & \multirow{2}{*}{0.455} \\
     & Load  & 88.6 & 28.2 & 15.6 & 63.4 & 24.7 & 52.2 & & & & & \\
\addlinespace[1pt]
GINO & Field & 6.3 & 10.2 & 18.0 & 25.6 & 20.3 & 29.4 & \multirow{2}{*}{0.254} & \multirow{2}{*}{1.298} & \multirow{2}{*}{0.216} & \multirow{2}{*}{0.569} & \multirow{2}{*}{0.539} \\
     & Load  & 31.0 & 8.5 & 3.7 & \textbf{$-13.1$} & 21.0 & \textbf{$-1.6$} & & & & & \\
\addlinespace[1pt]
Transolver++ & Field & 80.0 & 48.0 & 58.7 & 33.0 & 23.5 & 38.4 & \multirow{2}{*}{0.030} & \multirow{2}{*}{0.462} & \multirow{2}{*}{0.356} & \multirow{2}{*}{0.508} & \multirow{2}{*}{0.534} \\
             & Load  & 86.8 & 32.3 & 14.5 & 27.4 & 23.2 & 31.2 & & & & & \\
\addlinespace[1pt]
Unisolver & Field & 16.5 & 17.6 & 25.8 & 1.0 & 3.5 & 22.0 & \multirow{2}{*}{0.272} & \multirow{2}{*}{0.598} & \multirow{2}{*}{0.189} & \multirow{2}{*}{0.618} & \multirow{2}{*}{0.535} \\
          & Load  & 24.1 & 19.1 & 4.0 & 28.3 & 23.2 & 15.3 & & & & & \\
\addlinespace[1pt]
MPP & Field & 65.2 & 26.8 & 45.9 & 29.1 & 11.7 & 32.9 & \multirow{2}{*}{0.141} & \multirow{2}{*}{0.614} & \multirow{2}{*}{0.367} & \multirow{2}{*}{0.219} & \multirow{2}{*}{0.626} \\
    & Load  & 80.0 & 9.5 & 14.1 & 54.6 & 6.3 & 34.1 & & & & & \\
\addlinespace[1pt]
CALM-PDE & Field & 39.3 & 15.6 & 34.3 & \textbf{$-51.9$} & \textbf{$-10.0$} & 3.1 & \multirow{2}{*}{0.019} & \multirow{2}{*}{0.451} & \multirow{2}{*}{0.390} & \multirow{2}{*}{0.368} & \multirow{2}{*}{0.450} \\
         & Load  & 22.7 & 6.9 & 2.9 & \textbf{$-28.0$} & 17.2 & \textbf{$-12.8$} & & & & & \\
\addlinespace[1pt]
UPT & Field & 84.0 & 53.9 & 62.5 & 33.6 & 28.9 & 34.7 & \multirow{2}{*}{0.010} & \multirow{2}{*}{0.417} & \multirow{2}{*}{0.362} & \multirow{2}{*}{0.364} & \multirow{2}{*}{0.504} \\
    & Load  & 88.5 & 40.3 & 16.4 & 35.1 & 25.7 & 33.4 & & & & & \\
\addlinespace[1pt]
GEPS & Field & 54.9 & 17.3 & 37.4 & 24.2 & \textbf{$-2.9$} & 22.8 & \multirow{2}{*}{0.356} & \multirow{2}{*}{0.389} & \multirow{2}{*}{0.224} & \multirow{2}{*}{0.236} & \multirow{2}{*}{0.630} \\
     & Load  & 61.6 & 5.0 & 8.5 & 25.7 & 9.0 & 23.9 & & & & & \\
\bottomrule
\end{tabular*}
}
\caption{Stratified paired gains and route-level MCF on the predefined $5{,}004$-pair evaluation set. Positive gains indicate lower STCO error, bold marks negative gains, and ``External'' denotes external onset. STCO improves $68/72$ field cells and $63/72$ pressure-derived load cells, with mean gains of $31.1\%$ and $24.7\%$. For MCF, $\tau$ is replaced by another same-band lead, $\psi$ and $\Delta\psi$ are drawn from fixed channelwise Gaussian marginals, and $\bm g$ and $\bm u_{\mathrm{bc}}$ use active benchmark fields. Deterministic draws are shared across models; unchanged-input repeats define the numerical floor. GAOT has the largest mean field gain, and GINO the largest spatial MCF.}
\label{tab:supp-stratified-gains}
\end{table*}

\begin{figure*}[!t]
    \centering
    \includegraphics[width=0.97\textwidth]{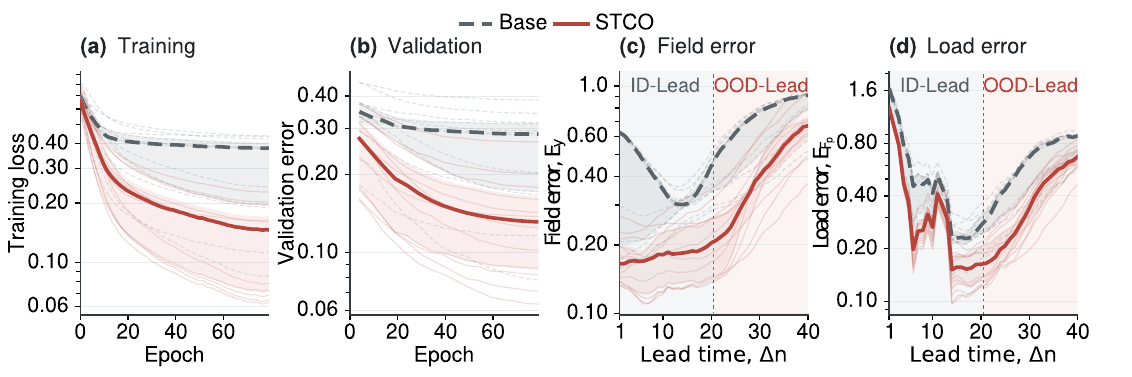}
    \caption{Learning and lead-time dynamics across twelve matched Base--STCO pairs.
    (a) Training objective and (b) validation relative-$L_2$ error over $80$ epochs.
    (c) Velocity--pressure field error $E_y$ and (d) pressure-derived load error $E_{F_p}$ over the predefined $5{,}004$-pair evaluation set.
    Thin curves show individual backbones; thick curves and bands show the cross-backbone median and interquartile range.
    Each lead is an independent target-time query. Gray and red backgrounds denote ID-Lead and OOD-Lead.
    Before smoothing, STCO lowers field error in $466/480$ and load error in $454/480$ backbone--lead comparisons.}
    \label{fig:supp-learning-dynamics}
\end{figure*}

\begin{figure*}[t]
    \centering
    \includegraphics[width=0.915\textwidth]{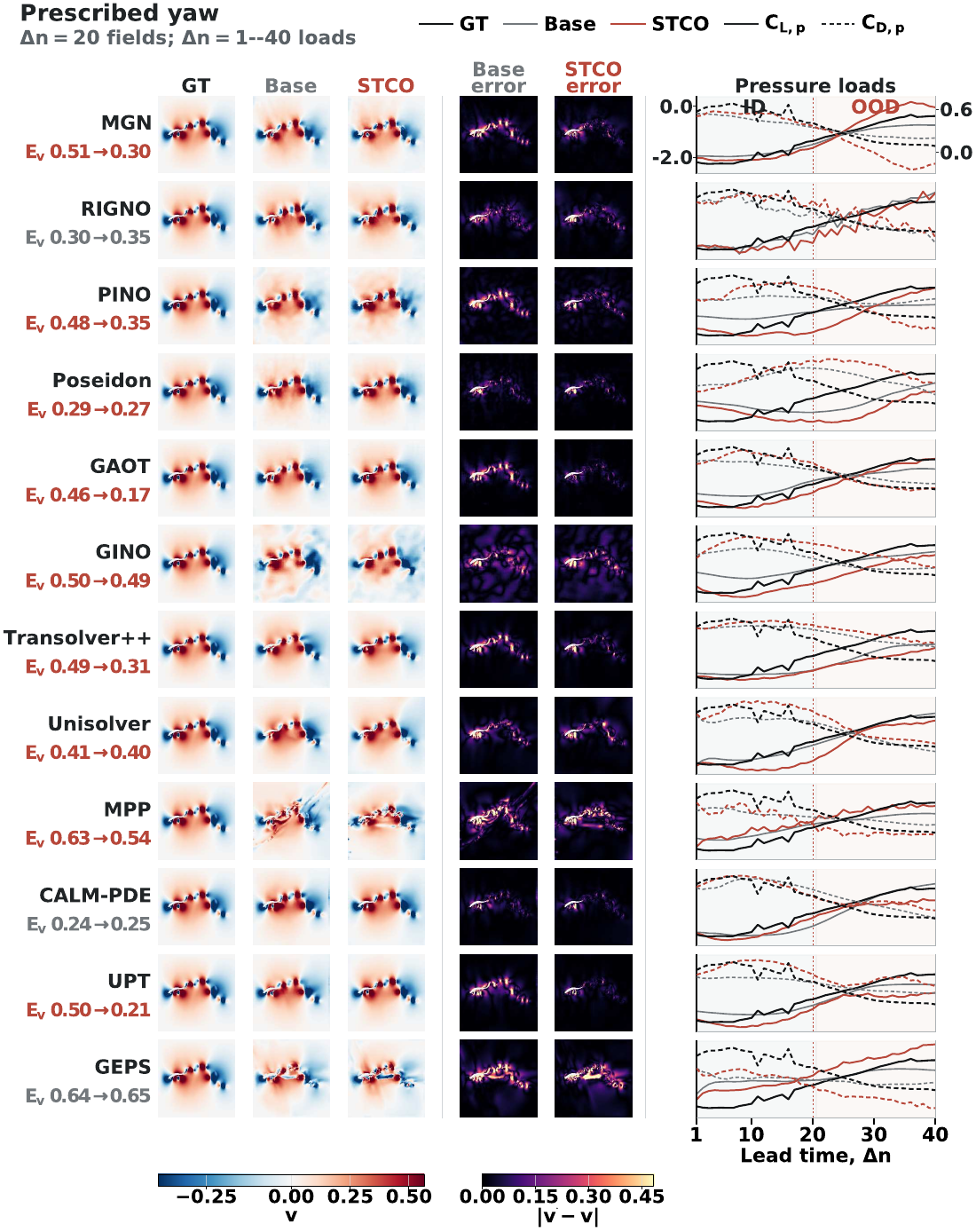}
    \caption{Prescribed-yaw response across twelve matched backbones.
    At $n_r=128$ and $\Delta n=20$, rows show GT, Base/STCO transverse velocity $v$, absolute errors, and fluid-region relative-$L_2$ error $E_v$, followed by pressure-derived lift (solid) and drag (dashed) over independent target-time queries at $\Delta n=1$--$40$.
    Field, error, and load-axis scales are shared across rows. Gray/red backgrounds mark ID/OOD-Lead.
    STCO lowers $E_v$ for $9/12$ backbones, with a mean reduction of $18.9\%$. GAOT and UPT show the largest reductions, $62.8\%$ and $57.5\%$.}
    \label{fig:supp-yaw-all-backbones}
\end{figure*}

\begin{figure*}[t]
    \centering
    \includegraphics[width=0.915\textwidth]{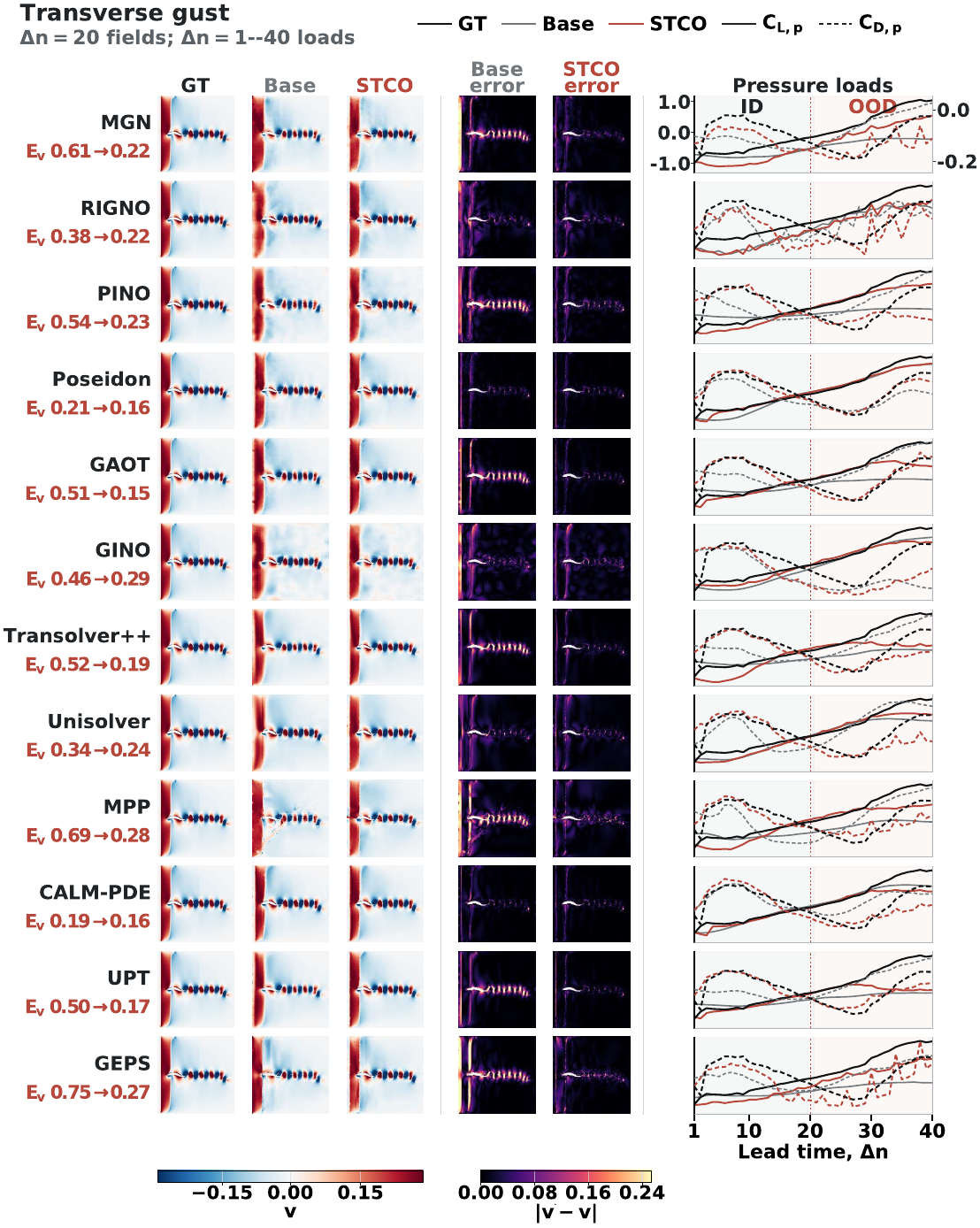}
    \caption{Transverse-gust response across twelve matched backbones at $n_r=126$ and $\Delta n=20$.
    Rows use the field and pressure-derived load layout of Figure~\ref{fig:supp-yaw-all-backbones}. The load curves span independent target-time queries at $\Delta n=1$--$40$.
    STCO lowers $E_v$ for all twelve backbones, with a median reduction of $58.5\%$.
    It lowers the pressure-derived load error of Equation~\ref{eq:supp-force-metric} over the forty leads for $11/12$ backbones, with a median reduction of $34.3\%$.}
    \label{fig:supp-tg-all-backbones}
\end{figure*}

\begin{figure*}[t]
    \centering
    \includegraphics[width=0.95\textwidth]{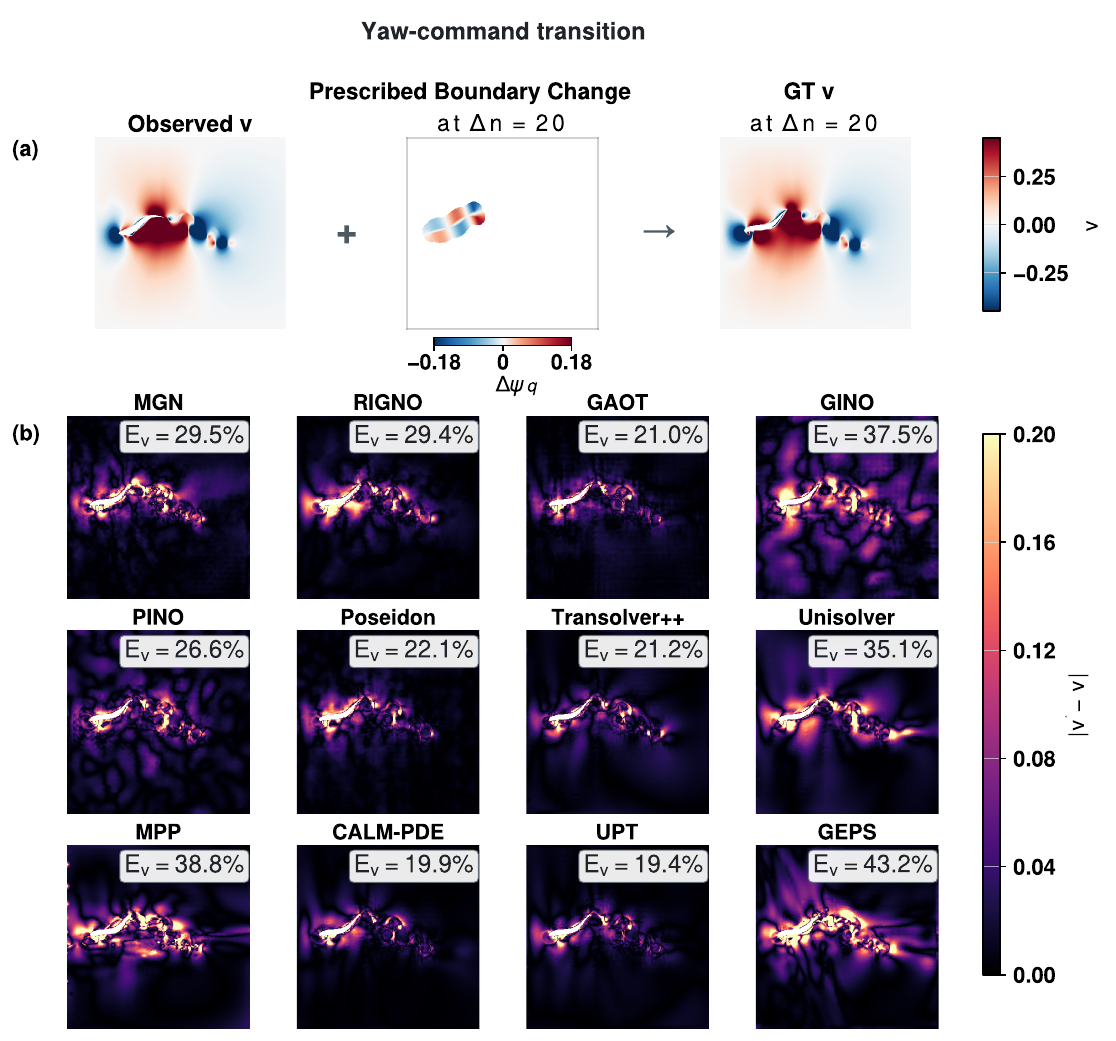}
    \caption{Prescribed-yaw transition across twelve STCO configurations.
    (a) Transverse velocity $v$ at the observed frame $n_r=44$, target-time signed-distance change $\Delta\psi_q$, and GT $v$ at $\Delta n=20$.
    (b) Pointwise absolute errors $|\widehat v-v|$. Labels report full-mesh fluid-region relative-$L_2$ error $E_v$.
    Observed and GT $v$ share one color limit, and all error panels share another.
    The median $E_v$ across the twelve models is $0.280$. Errors concentrate near the moving body and wake.}
    \label{fig:supp-yaw-transition}
\end{figure*}

\begin{figure*}[t]
    \centering
    \includegraphics[width=0.95\textwidth]{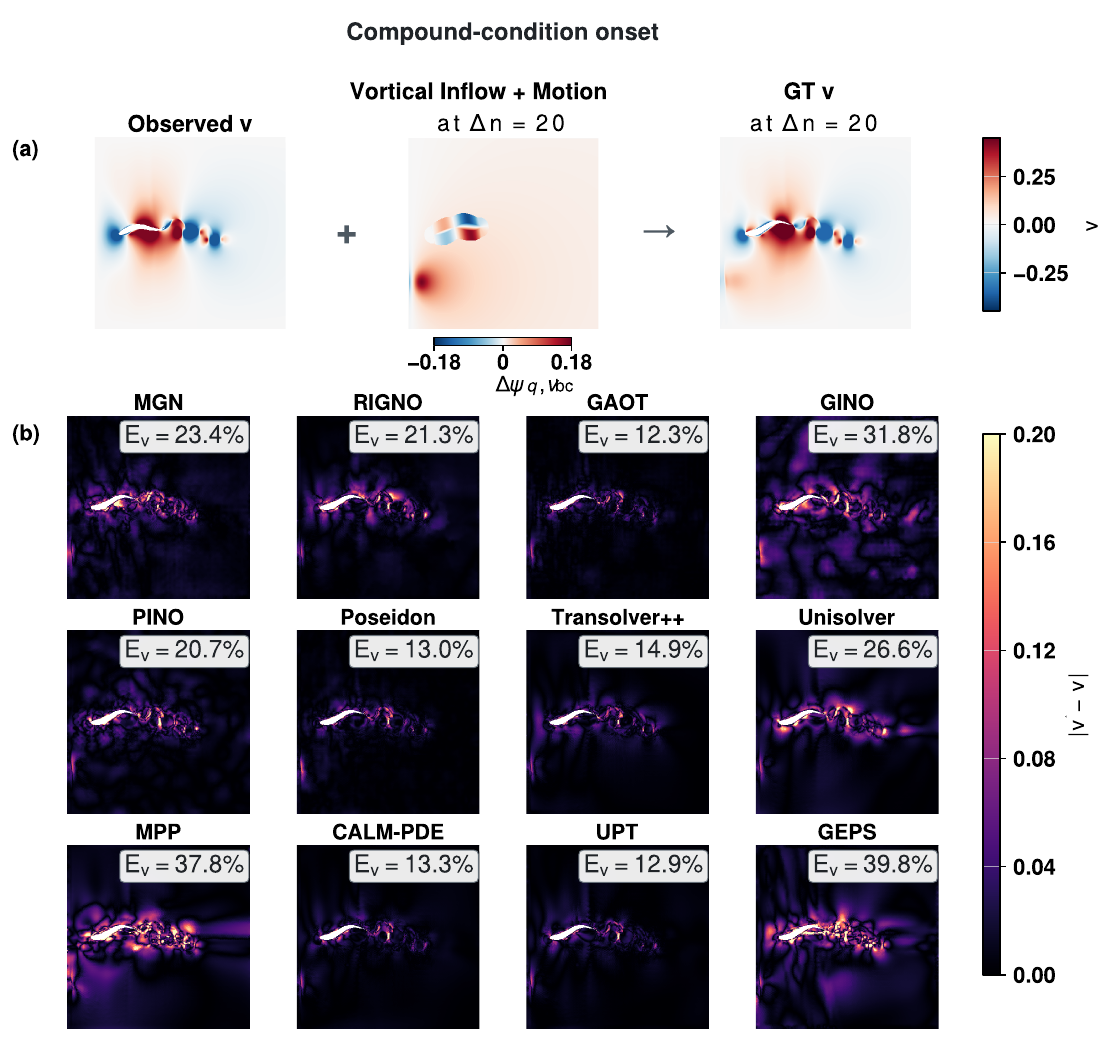}
    \caption{Compound vortical-inflow and motion response across twelve STCO configurations.
    (a) Transverse velocity $v$ at the observed frame $n_r=22$, target-time $v_{\mathrm{bc},q}$ overlaid with $\Delta\psi_q$, and GT $v$ at $\Delta n=20$. The two condition fields share the range $[-0.18,0.18]$.
    (b) Pointwise absolute errors $|\widehat v-v|$. Labels report full-mesh fluid-region relative-$L_2$ error $E_v$.
    The median $E_v$ across the twelve models is $0.210$. Errors concentrate near the moving body and disturbed wake.}
    \label{fig:supp-compound-vortical}
\end{figure*}

\FloatBarrier

\section*{Acknowledgments}

This research is funded by Engineering Start-up Grant of King’s College London,
and the Daiwa Anglo-Japanese Foundation through Daiwa Foundation Awards
(14465/15310).
The authors acknowledge the use of King's Computational Research,
Engineering and Technology Environment (CREATE) in conducting this
research~\cite{kclcreate}.

\end{document}